\documentclass[11pt]{article}
\usepackage{graphicx}
\usepackage{setspace}
\usepackage[style=nejm, articledoi=true, backend=biber, url=false, mincitenames=1, maxcitenames=2, minbibnames=1, maxbibnames=6]{biblatex}
\DeclareNameAlias{cite}{family}

\renewcommand{\cite}{\supercite}

\DeclareCiteCommand{\citet}
  {\usebibmacro{prenote}}
  {%
    \printnames{labelname}%
    \ifnumgreater{\value{listtotal}}{2}{\addspace\textit{et al.}}{}%
    \textsuperscript{\printfield{labelnumber}}%
  }
  {\multicitedelim}
  {\usebibmacro{postnote}}

\usepackage{pdflscape}
\usepackage{mathptmx}
\usepackage{xcolor}
\usepackage{placeins}
\usepackage{geometry}
\usepackage{xr-hyper}
\usepackage{hyperref}

\usepackage{comment}
\usepackage{booktabs}
\usepackage{multirow}
\usepackage{authblk}
\usepackage{rotating}
\usepackage{tabularx}
\usepackage{changepage}
\usepackage{makecell}
\usepackage{longtable}

\usepackage{minitoc}
\newcommand{\mdpt}{MDPT}
\newcommand{\mdp}{Medical Data Pecking}
\newcommand{\aou}{AoU}
\newcommand{\ttd}{T2D}
\newcommand{\ckd}{CKD}
\newcommand{\chf}{CHF}
\newcommand{\htn}{HTN}

\title{A Generative Approach for Semantic Auditing of Electronic Health Records}
\author[1]{Irena Girshovitz}
\author[2]{Atai Ambus}
\author[2]{Moni Shahar}
\author[1,3, $^{*}$]{Ran Gilad-Bachrach} 
\affil[1]{School of Biomedical Engineering, Faculty of Engineering, Tel Aviv University, Tel Aviv, Israel}
\affil[2]{AI and Data Science Center of Tel Aviv University (TAD), Tel Aviv, Israel}
\affil[3]{Safra Center for Bioinformatics, Tel Aviv University, Tel Aviv, Israel}
\affil[${*}$]{Corresponding author: rgb@tauex.tau.ac.il}
\date{}

\AtBeginBibliography{\setstretch{2}}

\begin{document}

\maketitle

\begin{abstract} 
The reliability of clinical artificial intelligence (AI) depends on high-quality data, yet Electronic Health Records are often inconsistent with existing scientific knowledge. Current quality assessments are limited: they either focus on syntax or rely on labor-intensive manual rules to capture semantic nuances. To overcome these scalability barriers, we propose Medical Data Pecking, a methodology that adopts software unit testing principles for medical data validation. It introduces Semantic Data Coverage, employing Large Language Models to generate context-aware tests that "peck" for inconsistencies between observed data and epidemiological evidence. To demonstrate this methodology, we implemented a reference tool using a Retrieval-Augmented Generation architecture that synthesizes medical literature into executable code. When applied to three datasets, this implementation generated dozens of tests per cohort, identifying discrepancies between observed distributions and epidemiological priors. These discrepancies encompass both genuine data inconsistencies and expected cohort-selection effects. This work provides an initial framework for scalable semantic auditing, shifting assurance from manual rules to the generative and context-sensitive verification required for trustworthy AI.

\end{abstract}

\section*{Introduction}
Electronic health records (EHRs) are increasingly used in healthcare research,\cite{prokosch_perspectives_2009, seminara_validity_2011, coorevits_electronic_2013, chen_research_2023} in developing machine learning models for clinical use\cite{rakers_availability_2024, cardozo_use_2022, gjesvik_artificial_2024} and in the analysis of patient populations, treatment outcomes, and healthcare trends.\cite{Syed2023Digital}

However, EHR data quality remained a concern. \citet{weiskopf2013methods} found most EHR-based studies relied on basic data-quality checks. In a recent survey, $92\%$ of AI practitioners reported experiencing data quality issues causing negative downstream effects.\cite{sambasivan_everyone_2021}
\citet{kilkenny2018data} emphasize that poor data quality undermines results, coining the phrase "Garbage in - Garbage out". Accordingly, data must be accurate, valid, complete, and available.\cite{kahn2015transparent,kilkenny2018data} 

In this work, we propose the adoption of quality assurance techniques from engineering disciplines to enhance process efficiency and ensure reliable and high-quality outcomes.\cite{mitra2016fundamentals}
Specifically, in software engineering, unit tests are typically written prior to or immediately following code implementation.\cite{ellims2006economics} Ideally, all execution paths are covered by unit tests. These local tests are designed to evaluate a specific segment of code to verify its correctness and are run frequently to identify potential bugs early and close to their origin. Consequently, automated test suites are commonly integrated into the Continuous Integration/Continuous Development (CI/CD) process.\cite{Shahin_continuous_2017} 

We argue that in data-driven applications, software engineering best practices should be applied not only to the codebase but also to the data itself, as errors in either can lead to system failures or misleading conclusions. Such practices should encompass all components of the application, including the data preprocessing pipeline, where issues may affect any variable, not just those detected by downstream statistical analyses. Faults in raw data or early-stage processing can propagate unnoticed and ultimately compromise research validity.

The evaluation of data quality should consider the context of the analysis with respect to populations, geographies, specific time frames, and medical conditions.\cite{reimer2019veracity} Existing quality assessment methods often neglect these contextual factors, focusing instead on syntactic completeness.\cite{lewis2023electronic} 

Although concordance with external datasets and prior knowledge could bridge these gaps, most current tools omit this due to the complexity of manually defining rules for every possible clinical scenario.\cite{lewis2023electronic} We propose using Large Language Models (LLMs) to automate the creation of such rules for a given use case.

Recent breakthroughs in natural language processing, particularly the introduction of LLMs, are reshaping multiple domains including medical data analysis. LLMs have been applied to automate data extraction,\cite{ntinopoulos_large_2025} recognize clinical phenotypes,\cite{alsentzer_zero-shot_2023, yang_enhancing_2023, tekumalla_towards_2024} curate concept sets,\cite{anand_utility_2024} explore datasets,\cite{chen_genspectrum_2023} predict clinical risk,\cite{Han_evaluation_2024} and manage time-sequenced events during data preprocessing.\cite{mcdermott_event_2023} 
\citet{narayan_can_2022} evaluated LLMs across a range of data-related tasks, demonstrating state-of-the-art performance in data imputation, integration and error detection, even without task-specific fine-tuning.

Building on these capabilities, we introduce the \mdp{} approach. We propose \mdp{} to add systematic semantic auditing to complement existing data validation tools. While \mdp{} can perform baseline structural checks, its primary utility is the identification of Semantic Gaps - instances where data maintains structural validity yet lacks epidemiological consistency. We demonstrate this approach through a reference implementation that utilizes LLMs to dynamically generate unit tests tailored to the target population and dataset. Analogously to selective pecking behavior in birds, \mdp{} scans structured health records to identify relevant data fields, flag inconsistencies, and exclude extraneous content. It applies the generated tests to the data, identifies potential issues, and reports untested fields to quantify the coverage of the test suite.

\section*{Results}
We applied \mdpt{} across multiple datasets to evaluate whether generative semantic testing can identify discrepancies beyond those captured by syntactic validation.

\subsection*{Validation of Automated Test Generation}
\begin{figure}
    \centering
    \includegraphics[width=0.5\linewidth]{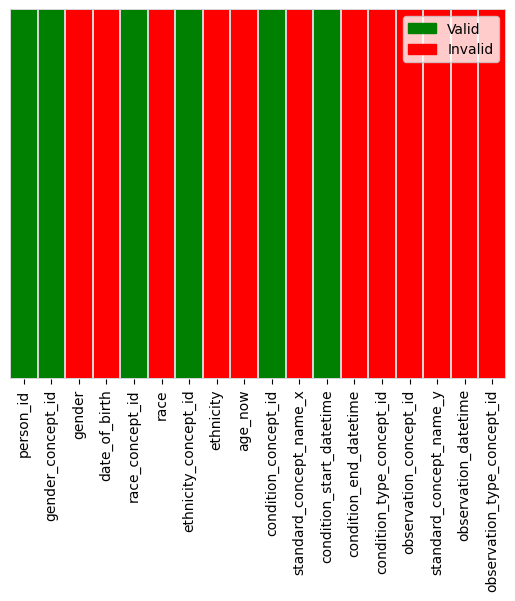}
    \caption{\textbf{Semantic Data Coverage Heatmap.} This representative example illustrates the density of semantic validation across clinical variables. The dense red clusters (see annotations) highlight systemic Semantic Gaps, areas where data is syntactically present (gray in standard tools) but fails validation against epidemiological priors (red in MDPT).}
    \label{fig:coverage_heatmap}
\end{figure}

To assess the reliability of the \mdp{} framework in synthesizing semantic unit tests, we evaluated the quality of the test suites produced by its reference implementation, the \mdp{} Tool (\mdpt{}), across four clinical cohorts: Type 2 Diabetes (\ttd{}) in All of Us (\aou{}), Chronic Kidney Disease in \aou{} (\ckd{}), Hypertension in SyntheticMass (\htn{}), and Congestive Heart Failure in MIMIC-III (\chf{}). The framework autonomously generated between 55 and 73 unit tests per cohort (an example is provided in Figure~\ref{fig:output_test}). The breadth of this coverage is visualized in Figure~\ref{fig:coverage_heatmap}, which displays the density of semantic validation across clinical variables, distinguishing between tested (green/red) and untested (gray) fields. 

Across all datasets, the Auditor Agent preserved test instances through automated correction rather than simple exclusion. In the \chf{} cohort, which presented the highest initial alignment variance, the Auditor corrected 55.0\% of the original tests to match clinical references, while only 10.0\% were discarded. For further quantification of the Auditor Agent’s role in refining this initial pool see Table~\ref{tab:auditor_correct_discard}. 
Following the verification pass, the final refined test suites across all cohorts are documented in Table~\ref{tab:all_testsֿֿֿֿ}.

\begin{figure}
    \centering
    \vspace*{-1.5cm}
    \includegraphics[width=0.8\linewidth, height=0.86\textheight, keepaspectratio]{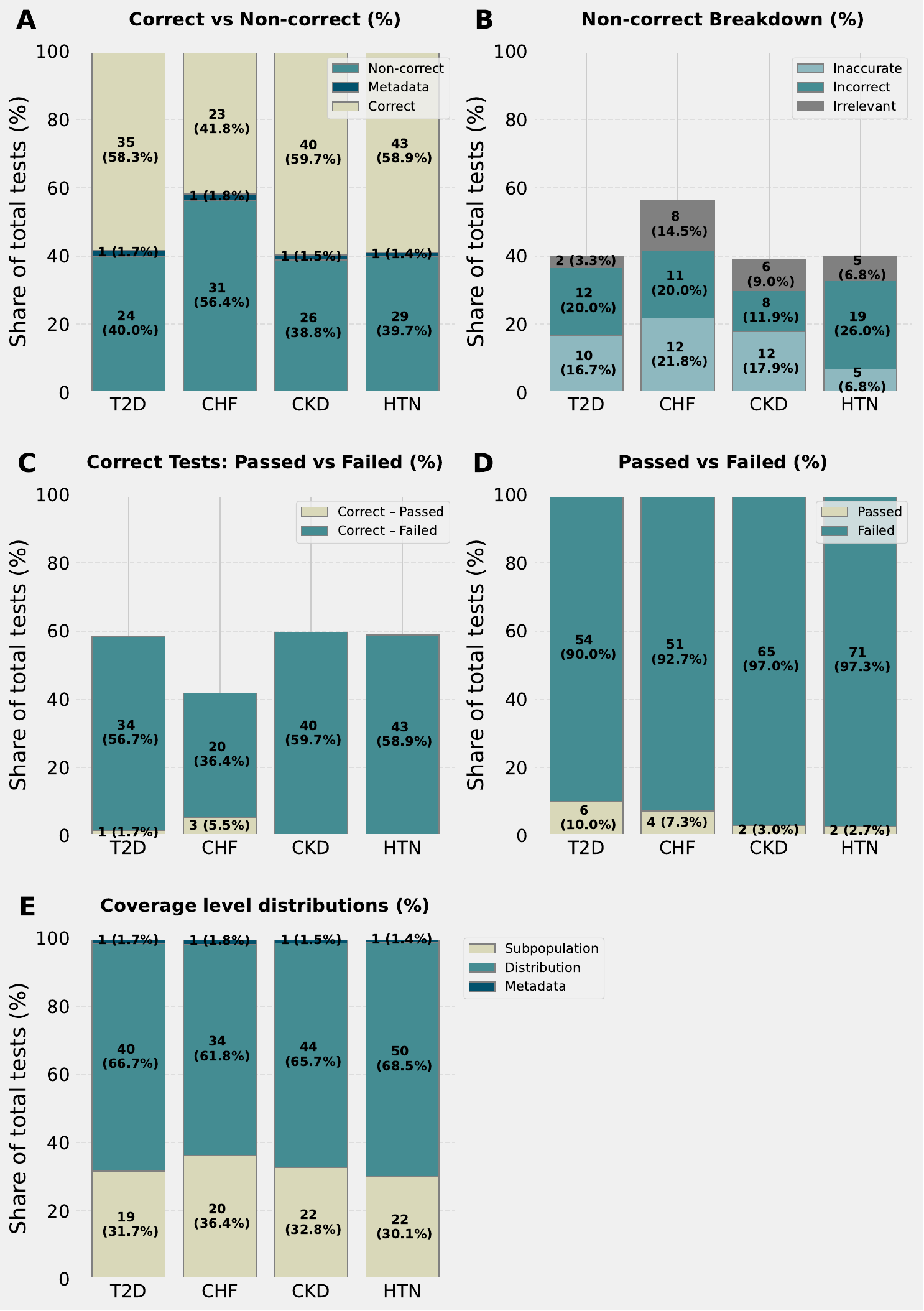}
    \caption{\textbf{Quantifying the Semantic Gap.} Evaluation summary of test suites generated by \mdpt{} across four cohorts (\ttd{}, \chf{}, \ckd{}, \htn{}).
Panels A–E show quality metrics normalized by the number of generated tests per cohort. Detailed definition of these metrics can be found in Section~\ref{sec:test_quality_val} in the Supplementary Materials:
(A) Proportion of tests with correct vs. non-correct reference values.
(B) Breakdown of non-correct test outcomes by subtype, stratified by pass/fail status.
(C) Distribution of correct test outcomes stratified by pass/fail status.
(D) Overall proportion of passed vs. failed tests.
(E) Distribution of tests across data coverage dimensions: subpopulation, distribution, and metadata levels.}
    \label{fig:results_all_datasets}
\end{figure}

We evaluated the correctness of the literature-derived expected values against the source text and other published and publicly accessible sources in a single-reviewer curation process.
In all suites, the reference validity ranged from 35\% to 93\% (Table~\ref{tab:test_quality}). The lower bounds were mainly driven by drug prevalence tests, for which public data on pharmaceutical usage is often unavailable or fragmented. When removing drug prevalence tests, the validity of the semantic anchors increased to 39–98\% (Table~\ref{tab:test_quality_no_drugs}), with accurate information grounding present in up to 69\% of the tests.

\subsection*{Sensitivity to Known Biases (Control Validation)}
Because real-world datasets lack a gold standard answer key for data quality, we validated the tool's discernment using a positive control approach.
To do this, we examined its performance on the \aou{} dataset, which is known to have specific recruitment skews.\cite{kathiresan2023representation}

Without manual configuration, the \mdpt{} framework correctly flagged demographic anomalies. Specifically, the generated semantic tests for gender and race distributions failed, accurately identifying the known over-representation of women and under-represented groups in the \aou{} cohort.\cite{kathiresan2023representation} This serves as a positive control, confirming that the dynamic semantic testing paradigm is capable of autonomously detecting deviations from general population priors without explicit user programming.

\subsection*{Uncovering Semantic Divergence}
The primary contribution of the semantic testing paradigm is the detection of discrepancies that pass syntactic checks (valid codes) but fail epidemiological checks (plausible meaning). The framework identified substantial divergence across all tested datasets (Table~\ref{tab:full_results}). Importantly, these divergences do not imply that the datasets are unsuitable for research. Instead, they highlight factors that should be considered when evaluating whether a dataset is appropriate for a specific research question. 

As detailed in Figure~\ref{fig:results_all_datasets}, our testing taxonomy is designed to capture semantic depth. Panel E illustrates that the vast majority of generated tests targeted the Distributional and Subpopulation (Contextual) semantic layers, while simple Metadata (syntactic) checks accounted for less than 2\% of the test suite. The consequences of this deeper validation are visible in Panel D: across all cohorts, between 90\% and 97\% of these semantic tests failed. 
Panel C reports outcomes restricted to the ‘Correct’ test subset. This subset consists of instances where the Auditor Agent successfully verified the literature-derived reference values against grounded epidemiological evidence. As shown in Panel C, failure rates within this verified subset were 87.1\% in the \chf{} cohort (36.4\% failed vs. 5.5\% passed), 97.3\% in the \ttd{} cohort (56.7\% failed vs. 1.7\% passed), and 100\% in both the \htn{} and \ckd{} cohorts.

In the \aou{} \ttd{} cohort, 54 of the 60 generated tests failed. Although the data were syntactically complete, the framework revealed substantial discrepancies between the observed prevalence of comorbidities and expectations derived from the literature. Similarly, in the \ckd{} cohort, 65 of the 67 tests failed.

In the MIMIC-III dataset evaluation for \chf{}, the framework detected a systematic formatting issue that rendered valid data semantically invisible. The tests failed due to a mismatch in ICD-9 coding standards (omission of decimal points). While standard syntactic checks validated these strings as non-null and allegedly correct, the semantic test, expecting a specific diagnosis prevalence, dropped to near zero, triggering a failure flag. This example shows that the framework can identify semantic discrepancies that may compromise downstream analysis. Although the omission of decimal points may appear to be a trivial formatting nuance, it represents a semantic break. Consequently, valid but dot-less ICD-9 codes were not captured by standard phenotype algorithms. \mdpt{} detected this by identifying that the observed prevalence of the disease was inconsistent with epidemiological priors.

\subsection*{Failure and Robustness Analysis}
Beyond distributional deviations, we examined failures in which the observed prevalence of a disease, measurement, or drug prescription was $0\%$. These cases were analyzed separately because a zero observed value may indicate something different from ordinary statistical misalignment. Depending on the context, it may reflect a technical artifact, a cohort specification gap, or a mismatch between the queried concept and the dataset representation. 

The pattern differed across cohorts. In the \chf{} cohort, zero-prevalence failures were primarily technical artifacts, mostly linked to the systematic omission of decimal points in ICD-9 codes. In the \htn{} cohort, these values were associated with gaps in the modeling of secondary comorbidities. In the \ckd{} and \ttd{} cohorts, they often reflected specification gaps during cohort construction, where relevant clinical items were overlooked in the initial cohort definition. 

Across all cohorts, medication-related zero-prevalence failures frequently resulted from granularity mismatch. The LLM-generated tests targeted highly specific medication codes that did not align with the ingredient-level abstraction or the modeling choices of the target datasets. A list of all zero-prevalence failed tests is provided in Table~\ref{tab:zero_results}.

\begin{figure}
    \centering
    \includegraphics[width=1\linewidth]{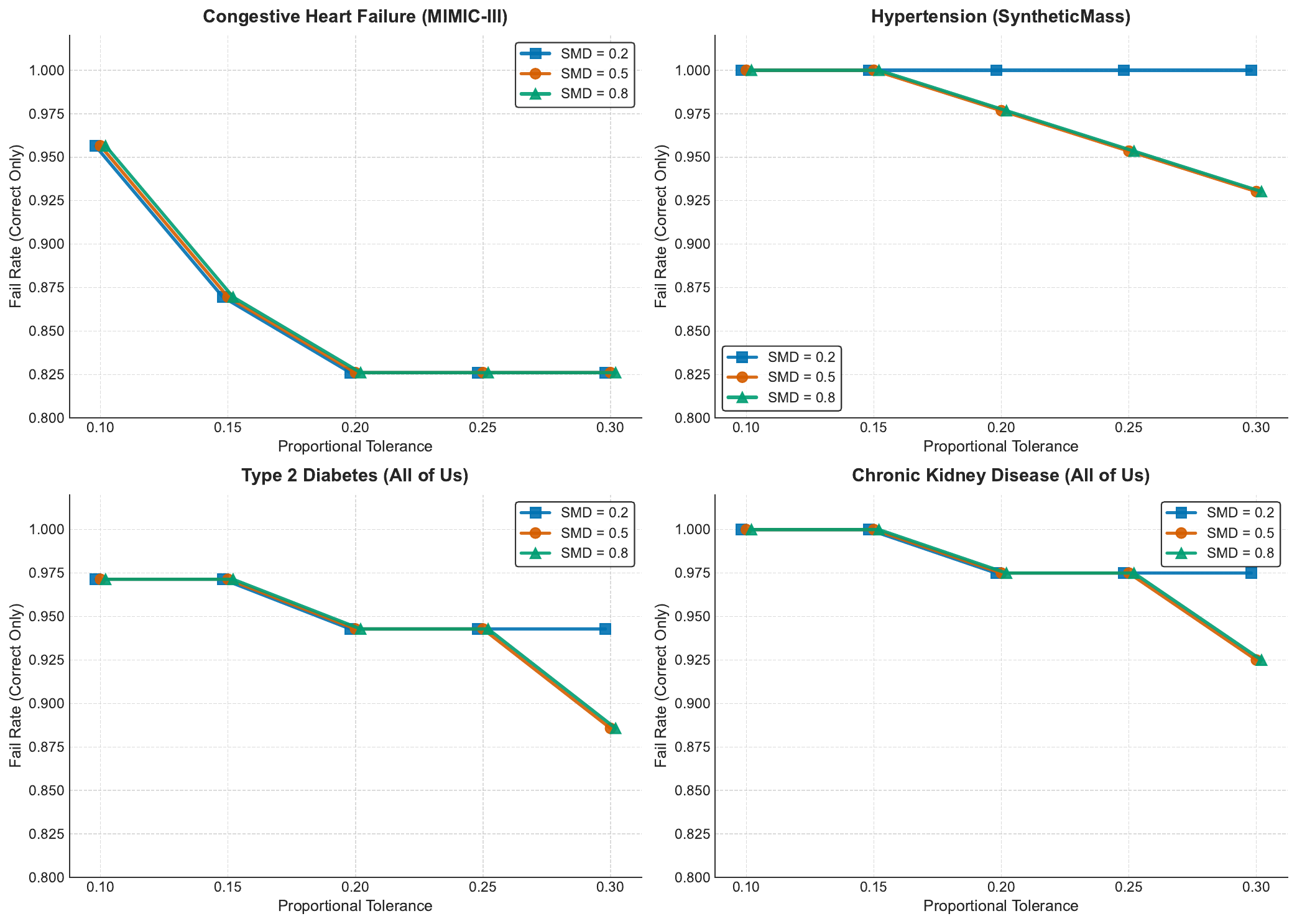}
    \caption{\textbf{Sensitivity analysis across statistical thresholds for correctly grounded tests.} Failure rates across correctly grounded tests are shown for proportional tolerance thresholds of 10\% to 30\% and SMD cutoffs of 0.2, 0.5, and 0.8. Across cohorts, failure rates varied more with proportional tolerance than with the SMD cutoff, and remained largely stable across SMD values.
}
    \label{fig:sensitivity_analysis_correct}
\end{figure}

We then evaluated the sensitivity of the framework to statistical threshold selection by varying the proportional tolerance threshold from 10\% to 30\% and the SMD threshold across 0.2, 0.5, and 0.8. As shown in Figure~\ref{fig:sensitivity_analysis_correct} for correctly grounded tests and in Figure~\ref{fig:sensitivity_analysis_all} for all tests, failure rates across all four cohorts were more sensitive to proportional tolerance than to the SMD threshold. By contrast, results remained largely stable across the evaluated SMD cutoffs, with only modest cohort-specific divergence at higher proportional tolerances. Full sensitivity values are provided in Table~\ref{tab:sensitivity_results}.

\section*{Discussion}
In this study, we introduced \mdp, a semantic unit testing methodology that adapts software engineering principles to medical data validation. While traditional quality assessment tools (such as OHDSI Achilles) provide extensive libraries of universal checks (e.g., 'Gender=Male cannot have pregnancy codes'), these rules are necessarily context-agnostic. A static rule library cannot capture the complexity of clinical context, such as the expected prevalence of a specific comorbidity in a distinct demographic cohort. Consequently, \mdpt{} functions as a complementary semantic validation layer. By addressing clinical implausibility in syntactically perfect data, the framework can identify discrepancies that may not be detected by traditional integrity checks. Using \mdpt{} we generated 23-43 semantic tests classified as correct and identified 20–43 discrepancies per cohort in data that were syntactically valid under standard checks. This demonstrates the benefit of using adaptive validation rather than a fixed set of rules. Our results suggest that syntactic completeness alone may be insufficient to establish semantic plausibility, and that verification against external evidence may reveal clinically relevant mismatches.

The high rate of discrepancies observed even among accurately grounded tests (Figure~\ref{fig:results_all_datasets}C) is consistent with a persistent semantic gap in these datasets. At the same time, these results should not be interpreted as evidence of data failure; rather, they show that \mdpt{} is able to bring these discrepancies to the attention of researchers.

Sensitivity analyses indicated that the main findings were more sensitive to proportional tolerance than to the SMD cutoff. Failure rates remained largely stable across the evaluated SMD values, suggesting that proportional tolerance had the greater influence on evaluation strictness.

Consequently, it is crucial to distinguish between data errors and cohort selection biases. A high failure rate in the \mdpt{} suite does not always imply corrupted data; rather, it quantifies the distance between the specific dataset and the general population baseline as reported in the scientific literature. For example, the high failure rate in MIMIC-III reflects its specific ICU population, which naturally deviates from outpatient epidemiological priors. 
Similarly, the failures observed in the SyntheticMass dataset may reflect the specific generative methodology of the Synthea engine, where certain complex clinical correlations may not have been modeled, rather than a failure of the data pipeline itself. In this setting, \mdpt{} can help characterize representational discrepancies introduced during synthetic data generation and identify how the resulting study population differs from the general population.

The detection of formatting-induced errors in MIMIC-III highlights how clinically meaningful information can become unavailable to downstream analyses despite remaining syntactically valid in the raw data. Although the underlying code strings are present and pass syntactic checks, the omission of a decimal point can alter the encoded signal enough to make the phenotype effectively undetectable by standard algorithms. This example shows that semantic testing can identify failures that conventional syntactic checks may miss through implausible prevalence estimates. In particular, zero-prevalence failures provide a distinct diagnostic signal that suggests either a technical artifact or a mismatch between the clinical query and the level of abstraction represented in the dataset. By identifying these cases, the framework helps distinguish routine distributional differences from more fundamental incompatibilities between the intended research question and the structure of the available data.

The imperative for this dynamic validation is growing. The White House Office of the National Cyber Director advocated the development of quality metrics to evaluate software modularity, security, and automation.\cite{the_white_house_back_2024} Similarly, the Bipartisan House AI Task Force identified data quality as a critical concern.\cite{noauthor_bipartisan_2024} \citet{alderman_tackling_2025} reinforced these concerns through the STANDING Together recommendations, a global initiative to mitigate algorithmic bias in healthcare AI. In the era of Big Data, manual verification is infeasible and errors in the data pipeline propagate unnoticed, undermining the validity of downstream AI models. Our framework provides a scalable and automated approach for addressing this validity gap, offering a mechanism to operationalize these high-level policy and ethical recommendations into code.

A primary challenge in applying LLMs to quality assurance is the risk of hallucination or faulty tests. We addressed this through a tiered evidence grounding protocol reinforced by an automated verification loop. Rather than relying on the LLM's knowledge, our system functions as a retrieval engine, anchoring every test in verifiable sources. Crucially, the system utilizes a recursive search strategy: after the initial generation, a secondary 'Auditor Agent' independently searches for and verifies the statistics, allowing the system to self-correct and discard unsupported claims before they become unit tests. Although our evaluation showed that drug prevalence tests remain difficult to ground due to data scarcity, the stronger grounding of disease and demographic tests suggests that RAG-based architectures with independent verification cycles may support automated epidemiological validation.

In this study, we use an auditor agent to reduce major grounding errors. Nevertheless, there is still room for improvement that may be achieved by using more advanced LLMs and by conducting additional iterations of "debate" between agents until they agree on the values reported in the document.

To maintain clinical reliability, the \mdp{} framework separates generative synthesis of tests from their subsequent execution. By fixing the validated test matrix as a versioned asset within the clinical pipeline, the system provides a stable audit trail. This architecture mitigates reproducibility risks associated with evolving external APIs and helps maintain consistency of the audit over time.

As \citet{dahl1972structured} argued, testing can be used to show the presence of bugs, but never to show their absence. Our current implementation is limited by its reliance on general population benchmarks, which can obscure deviations in specific subgroups. However, this is largely a limitation of data availability rather than of the methodology itself. 

While \mdpt{} focuses on prevalence, the framework’s architecture can be expanded to support deterministic semantic tests, such as temporal coherence and dosage plausibility.

Moreover, the reliance on general population priors introduces a form of 'reverse ecological fallacy': the assumption that a specialized, highly filtered clinical cohort (such as an ICU population) should mirror broad national averages. This reliance creates an inherent noise floor, where \mdpt{} may flag valid clinical divergence, necessary for the study of specific diseases, as a semantic error.

While temporal mismatch is a systemic challenge, it is the representational discrepancy, specifically formatting-induced errors in clinical coding, that serves as the primary driver for failures in specific phenotypes like \chf{}. In MIMIC-III, the omission of decimal points in ICD-9 codes prevents automated phenotype algorithms from correctly identifying conditions, creating a semantic gap that likely exceeds the noise introduced by shifting epidemiological priors. However, even with perfect formatting, temporal mismatch remains a critical secondary factor when analyzing historical cohorts. Comparing decade-old clinical signals against contemporary benchmarks introduces an 'era gap' in diagnostic criteria and prevalence. Thus, while the immediate hurdle is technical (ICD-9 formatting), era-aware grounding is essential for the longitudinal validity of generative semantic testing.

As local health systems digitize their internal baselines, these can replace the general literature as the ground truth. Furthermore, the impact on the validity of subsequent research must be carefully considered; there is a risk of the \emph{implied truth effect} where researchers can interpret the absence of warnings as an indication of validity.\cite{pennycook2020implied} Semantic testing is designed to detect known impossibilities, not to certify absolute truth. To mitigate it, we propose that generative auditing must adopt a tri-state reporting protocol where clinical variables are explicitly categorized as Passed, Failed (Semantic Gap), or Inconclusive (No Reference or Out-of-Scope). By treating 'No Reference' results as a distinct category rather than a default pass, the system ensures that researchers interpret an absence of a flag as a lack of evidence rather than an endorsement of data quality. This transparency may help reduce over-interpretation and support safer integration of \mdp{} into clinical data governance. 

It is important to distinguish the methodology (semantic unit testing) from the mechanism (LLMs). While our current implementation relies on GPT-4o and Bing Search, the framework itself is model-agnostic. As foundational models evolve, the 'Auditor Agent' can be upgraded to specialized biomedical models, but the underlying logic, which is validating data against dynamic external truth, remains the main methodological contribution of this work.

Ultimately, these findings support more routine use of automated data testing pipelines in medical AI. Similarly to modern software development where automated testing pipelines are standard, medical AI may benefit from comparable testing workflows. Furthermore, using such tests continuously on data collected by healthcare institutes can help identify problems at the source.

\FloatBarrier

\section*{Methods}
The \mdp{} framework consists of three main components: (i) a taxonomy of semantic unit tests, (ii) automated methods for test synthesis, and (iii) procedures for applying these tests to data. This section describes each component in turn and then presents the datasets used to evaluate \mdpt{}, the reference implementation of the framework.

\subsection*{Taxonomy of Semantic Unit Tests}
To systematically evaluate the extent to which data were validated for research, we propose a hierarchical taxonomy of data validation based on the software engineering principle of "test coverage".\cite{kochhar_code_2015} Traditional quality metrics primarily ensure syntactic coverage, verifying that fields adhere to correct formats. While some frameworks extend this to semantic validation, they rely on manually curated rules to detect logical errors. This approach is labor-intensive and not scalable to the complexity of medical data. In contrast, our approach introduces Semantic Data Coverage to validate the alignment of observed data with epidemiological ground truth. Specifically, we distinguish three distinct levels of validation: the syntactic (metadata) level, the distributional semantic level, and the contextual semantic (subpopulation) level.

    Syntactic level (Metadata): Tests at this level verify structural integrity, such as field presence, data type correctness, and non-missing values. This serves as a baseline validation to ensure that the data are technically accessible.

    Distributional semantic level: Tests at this level validate that population metrics align with theoretical or literature baselines. For example, the distribution of a specific laboratory test across the entire cohort is checked against its expected range. This ensures that the data are statistically plausible in aggregate.

    Contextual semantic level (Subpopulation): Tests at this level analyze intra-group patterns, such as comorbidity rates or drug prevalence within specific demographic cohorts. For example, these tests might check the prevalence of hypertension in women compared to men. This level offers greater depth than population checks by detecting context-specific biases or errors that remain invisible to aggregate analysis.

\subsection*{Core Concepts and Framework Overview}
For clarity, we use the following terms throughout the framework:

    Semantic Unit Testing: A methodology that adapts software engineering unit testing principles to validate discrete clinical segments of medical data against external scientific knowledge.
    
    Semantic Data Coverage: The extent to which observed data fields are validated for alignment with epidemiological ground truth, moving beyond simple formatting checks.
    
    Semantic Gap: A statistically significant divergence where data maintains Syntactic Integrity (adherence to format and metadata) but lacks Semantic Plausibility (alignment with clinical reality).

A schematic description of the semantic testing depth hierarchy is presented in Figure~\ref{fig:MDPT_schematic}a. To implement this semantic taxonomy at scale, we developed the \mdpt{}. As illustrated in Figure~\ref{fig:MDPT_schematic}b, the framework utilizes an LLM to automatically synthesize these tests from the medical literature and run them against the target dataset.

\begin{figure}[t]
    \centering
    \includegraphics[width=\textwidth]{Fig1.pdf}
    \caption{\textbf{The Semantic Unit Testing Paradigm and \mdpt{} Architecture.} 
    (a) The Semantic Gap in data quality. Standard quality assessment tools (red) typically provide only syntactic coverage (metadata and formatting). The proposed framework extends validation to the distributional (yellow) and contextual (green) semantic layers to detect errors in meaning. 
    (b) The generative workflow. The system operates in two stages. In the first stage (left panel), it retrieves epidemiological statistics via web search (Bing) while simultaneously mapping clinical concepts to standard vocabularies via vector search. In the second stage (right panel), it synthesizes these inputs into a test matrix and performs 'Expected Value Validation', a double-check by an Auditor Agent to prevent hallucinations before generating the final unit tests.}
    \label{fig:MDPT_schematic}
\end{figure}

\begin{figure}
    \centering
    \includegraphics[width=1\linewidth]{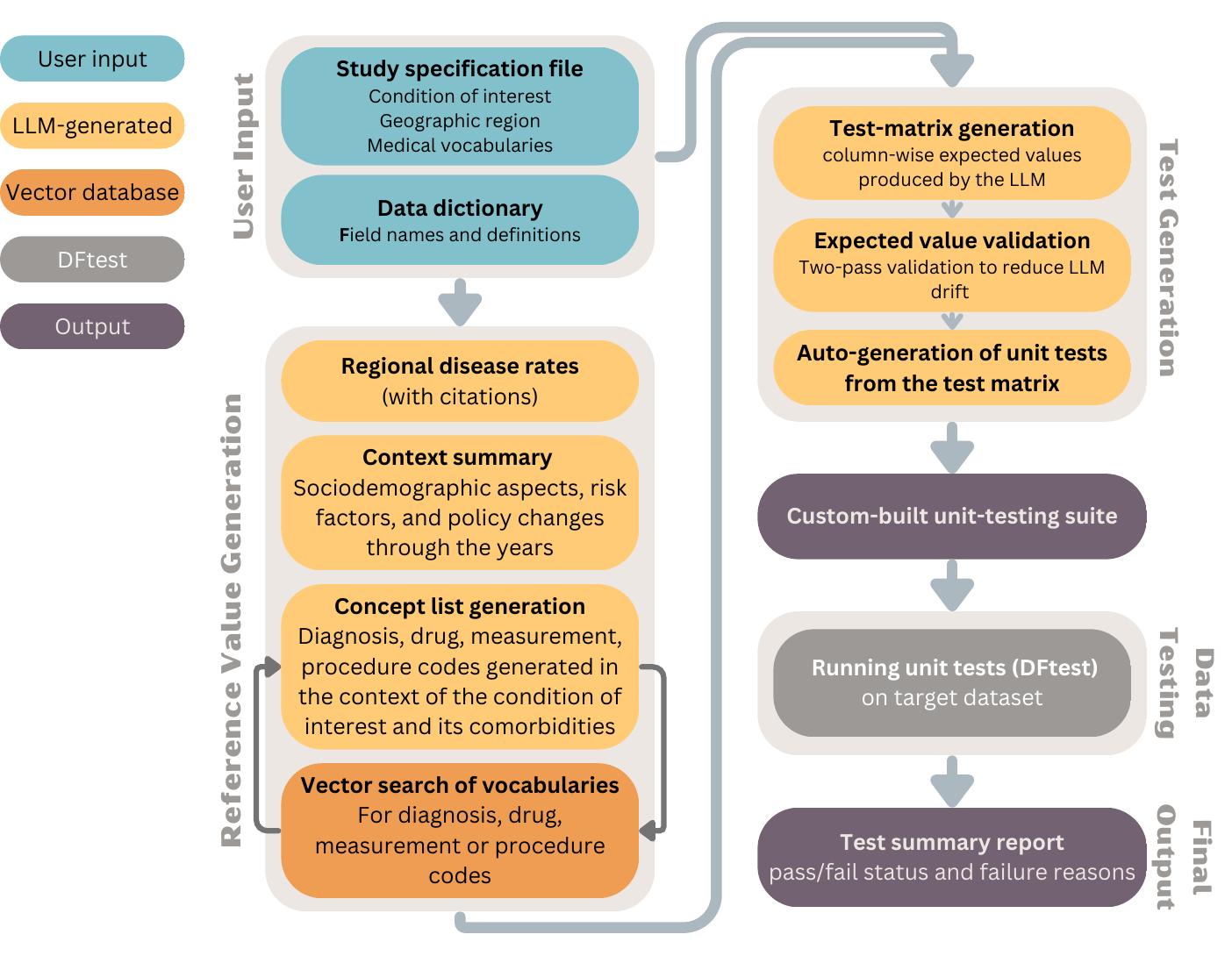}
    \caption{\textbf{A schematic overview of the \mdpt{} workflow.}
    User-defined study parameters are input into an LLM module that retrieves regional disease rates, contextual epidemiology, and generates lists of diagnosis, drug, measurement, and procedure codes via vector-database search. These reference values are transformed into a test matrix, implemented as unit tests, and validated in a double-pass step to reduce LLM hallucinations. The DFtest framework then executes the suite on the target dataset, yielding binary test outcomes and statistical comparisons between observed data and expected epidemiological distributions.}
    \label{fig:MDPT_workflow}
\end{figure}

\subsection*{Automated Test Synthesis}
To support deployment in secure clinical environments, the \mdpt{} framework utilizes a decoupled architecture that separates logic generation from data execution. The system is organized into two functionally isolated modules: the Generation Module and the Test Application Module.

Generation Module: This module operates exclusively on the study specification and the data dictionary, which describes the available database fields, their meanings, formats, and coding. The LLM utilizes this metadata to map clinical intent to the local database schema and generate executable unit tests. No raw, individual-level patient data is ever transmitted to or accessed by this module.

Test Application Module: This module resides within the user’s secure, isolated environment. It receives the generated test suite and executes it directly against the raw patient data. This separation of concerns ensures that protected health information remains within the local security perimeter while utilizing generative models for test construction.

\subsection*{Contextual Parameter Extraction and Grounding}
To address the scalability challenge of creating context-specific unit tests, the \mdpt{} framework employs a Retrieval-Augmented Generation (RAG) architecture. The test generation pipeline operates in two primary stages: Contextual Parameter Extraction and Automated Grounding and Evidence Verification.

Contextual Parameter Extraction: The system accepts a high-level study specification (condition of interest, geographic region, and the medical ontology used) and a data dictionary (Figure~\ref{fig:json_def_example}). Using an LLM, the system identifies relevant medical concepts using a search engine API and aligns them using OHDSI standardized vocabularies (e.g., OMOP, ICD-9, SNOMED). Rather than performing a traditional source-to-concept mapping, the system utilizes these ontologies as a semantic reference library to define the clinical ground truth for the audit. Full model specifications and prompt templates are detailed in Section~\ref{supsec:specs} in the Supplementary Materials and Tables~\ref{tab:test_suggestion_prompts}-\ref{tab:test_val_prompts}. The specific vocabularies used in \mdpt{} are detailed in Table~\ref{tab:vector_vocabs}.
To ensure accurate concept mapping, standardized vocabularies are pre-embedded using OpenAI's "text-embedding-3-small" model and stored in a vector database (Chroma). To facilitate concept alignment, a structured metadata-augmented retrieval strategy was employed. For each identified clinical entity, a query string was constructed incorporating the clinical term, domain identifier, and vocabulary source (e.g., 'concept\_name: [diagnosis or comorbidity] domain\_id: [domain] vocabulary\_id: [user provided medical ontology] standard\_concept: S'). The system retrieves the top 50 candidate concepts from the Chroma vector database using cosine similarity. From this candidate pool, the LLM performs a secondary semantic verification to select the most appropriate standardized identifiers for the test suite.

\begin{table}
    \centering
    \small
    \begin{tabular}{p{2.75cm}p{5cm}p{4cm}l}
        \toprule
Vocabulary ID & Vocabulary Name & Vocabulary Version & Concept Count \\
\midrule
ICD10CM & International Classification of Diseases, Tenth Revision, Clinical Modification (NCHS) & ICD10CM FY2025 code descriptions & 99421 \\
\midrule
ICD9CM & International Classification of Diseases, Ninth Revision, Clinical Modification, Volume 1 and 2 (NCHS) & ICD9CM v32 master descriptions & 17564 \\
\midrule
LOINC & Logical Observation Identifiers Names and Codes (Regenstrief Institute) & LOINC 2.77 & 271441 \\
\midrule
ICD9Proc & International Classification of Diseases, Ninth Revision, Clinical Modification, Volume 3 (NCHS) & ICD9CM v32 master descriptions & 4657 \\
\midrule
Ethnicity & OMOP Ethnicity & NA & 2 \\
\midrule
Race & Race and Ethnicity Code Set (USBC) & Version 1.0 & 53 \\
\midrule
Gender & OMOP Gender & NA & 5 \\
\midrule
ATC & WHO Anatomic Therapeutic Chemical Classification & ATC 2024-07-03 & 7131 \\
\midrule
DRG & Diagnosis-related group (CMS) & 2011-18-02 & 1362 \\
\midrule
RxNorm Extension & OMOP RxNorm Extension & RxNorm Extension 20240701 & 2146945 \\
\midrule
PPI & AllOfUs\_PPI (Columbia) & Codebook Version 0.4.43 + COVID + MHWB + SDOH + PFH + BHP + EHH & 6798 \\
\midrule
CPT4 & Current Procedural Terminology version 4 (AMA) & 2024 Release & 17455 \\
\midrule
ICD10 & International Classification of Diseases, Tenth Revision (WHO) & 2021 Release & 16638 \\
\midrule
SNOMED & Systematic Nomenclature of Medicine - Clinical Terms (IHTSDO) & 2024-02-01 SNOMED CT International Edition; 2024-03-01 SNOMED CT US Edition; 2024-04-10 SNOMED CT UK Edition & 1084286 \\
\bottomrule
    \end{tabular}
    \caption{\textbf{Clinical vocabularies included in the vector database used by \mdpt{} for source code retrieval}. The table lists each vocabulary identifier, its full name, its version used, and the total number of indexed concepts.}
    \label{tab:vector_vocabs}
\end{table}

Automated Grounding and Evidence Verification: To ensure that the tests are anchored in epidemiological reality, we implemented a three-step grounding protocol, which is illustrated in Figure~\ref{fig:MDPT_workflow}, comprising directed prioritization, test matrix construction, and Auditor Agent verification.

    Directed Prioritization: The system retrieves initial reference statistics via the Bing Search API. Through system prompting, the model is explicitly required to prioritize high-authority sources, such as government health agencies (e.g., CDC, NIH) and peer-reviewed literature, over general web results when selecting benchmarks.

    Test Matrix Construction: The retrieved epidemiological benchmarks are transformed into a structured test matrix. This matrix defines the preliminary expected value ranges for each semantic test (e.g., "The prevalence of Type 2 Diabetes in US adults is expected to be between 8\% and 12\%"). This automated synthesis allows the framework to generate dozens of context-aware tests from a single high-level query.

    The Auditor Agent (Double-Pass Verification): To mitigate hallucination, we used an approach inspired by Self-Refine,\cite{madaan2023self} where an "Auditor Agent" performs a new, targeted search to verify the accuracy of the statistics proposed in the matrix. This agent is explicitly required to validate the correctness of the extracted value; it cross-references the matrix entries against newly retrieved evidence and automatically corrects the value if discrepancies are detected. In this implementation, both the Generator and Auditor utilize GPT-4o; however, independence is maintained through information decoupling, where the Auditor initiates an entirely separate search thread, and role-specific prompting. This architecture ensures that the verification pass is grounded in a fresh evidentiary context.

\subsection*{Execution and Statistical Validation}
Unlike software unit tests, which typically require binary pass/fail outcomes, semantic data testing requires statistical tolerance to account for natural population variance and sample size differences. 
Adopting Software Engineering principles, we define a "test failure" as any objective non-alignment between a clinical assertion and the observed data. This technical definition is agnostic to the root cause; a failure flag may be triggered by technical data inconsistencies (e.g., formatting), scientifically valid cohort-selection effects (e.g., an ICU-specific design), or tool-related discrepancies.
We implemented a custom execution framework, DFtest, which applies the generated test matrix to the target dataset. While the synthesis of this matrix involves stochastic API processes, the resulting suite is persisted as a static, version-controlled asset. This decoupling ensures that the execution phase remains unaffected by updates or drift in the underlying commercial models or search algorithms. The framework applies the generated test matrix to the target dataset using three statistical criteria. These values were used as reference calibration settings in the baseline analysis and were not intended to represent universally optimal thresholds across all datasets or variable types:

Distributional Equivalence: For continuous variables (e.g., age at diagnosis), the framework applies Welch's t-test for unequal variances. A p-value less than 0.05 flags a statistically significant deviation from the expected distribution.

Magnitude of Difference: To distinguish between statistical significance and clinical relevance,\cite{van_rijn_statistical_2017} particularly in large datasets where minor deviations may yield low p-values, we calculate the Standardized Mean Difference (SMD). Discrepancies are flagged only when the SMD equals or exceeds 0.2, representing a non-negligible deviation from the expected distribution.

Proportional Tolerance: For categorical prevalence (e.g., comorbidity rates), we define a tolerance margin based on prior EHR validation studies.\cite{tatem_comparing_2017} A test passes if the ratio between the observed and expected proportion falls within the interval of 0.85 to 1.15. Ratios outside this range indicate a deviation of greater than 15\% from the expected baseline and trigger a failure flag.

\subsection*{Evaluation Datasets}
\label{sec:evaluationDatasets}
To demonstrate the generalizability of this paradigm, we deployed the framework across three datasets representing distinct data structures and coding systems.

    All of Us v7 (controlled tier CDR), from the All of Us research program (\aou{}; \url{https://www.researchallofus.org/}). A large-scale U.S. biobank targeting one million participants, with emphasis on historically underrepresented populations (OMOP). Population characteristics are described in Table~\ref{tab:TableOne_AoU}.
    
    MIMIC-III v.1.4,\cite{johnson_mimic-iii_2016} a publicly available de-identified critical care database of over 60,000 admissions to Beth Israel Deaconess Medical Center (ICD-9).
    
    SyntheticMass,\cite{walonoski_synthea_2017} a synthetic population dataset simulating the health records of Massachusetts residents (SNOMED-CT).

\section*{Data Availability}
The datasets analyzed during the current study are available under different access conditions. All of Us data are not publicly available due to controlled-access restrictions designed to protect participant privacy and are available to eligible researchers through the All of Us Researcher Workbench (\url{https://www.researchallofus.org/}). MIMIC-III is available to credentialed users through PhysioNet (\url{https://mimic.mit.edu/}). SyntheticMass is publicly available at \url{https://syntheticmass.mitre.org/download.html}.

\section*{Code availability}
The code used in this study is publicly available on GitHub at \url{https://github.com/TAU-MLwell/mdpt}. The implementation was developed in Python, and details of the software environment and dependencies are provided in the repository.

\section*{Acknowledgments}
We sincerely thank the participants of the \aou{} program for their invaluable contributions. We also acknowledge the National Institutes of Health's \href{https://allofus.nih.gov/}{All of Us Research Program} for providing access to the participant and cohort data used in this study.

The study was supported by the Center for AI and Data Science (TAD) at Tel-Aviv University

\section*{Author Contributions}
I.G. and R.G.B. conceived and designed the study and wrote the original draft of the manuscript. I.G., A.A., and R.G.B. developed the software framework. I.G. performed the data analysis, data curation, and visualization. R.G.B. and M.S. provided supervision and guidance throughout the project. All authors reviewed, edited, and approved the final manuscript.
 
\section*{Competing interests}
Ran Gilad-Bachrach is an employee of Microsoft. The company had no involvement in the study. All other authors declare no competing interests.

\FloatBarrier

\printbibliography

\clearpage
\appendix
\definecolor{grassgreen}{RGB}{50,170,50}

\newcommand{\highlightFail}[1]{\textcolor{red}{#1}}
\newcommand{\highlightPass}[1]{\textcolor{grassgreen}{#1}}
\newcommand{\highlightKeep}[1]{\textcolor{black}{#1}}
\newcommand{\highlightNoRef}[1]{\textcolor{orange}{#1}}

\title{Type 2 diabetes in US - Data Quality Evaluation Report}

\maketitle

\section{Demography and Diagnosis Tests}

\begin{longtable}{p{7.5cm} c c c c c}
\label{tab:demo_diagnosis}\\
\textbf{Test name} & \textbf{Result} & \textbf{Expected value} & \textbf{Actual value} & \textbf{SMD} & \textbf{ratio}\\
\hline
\endfirsthead
\multicolumn{6}{c}{{\bfseries Continued from previous page}}\\
\hline
\textbf{Test name} & \textbf{Result} & \textbf{Expected value} & \textbf{Actual value} & \textbf{SMD} & \textbf{ratio}\\
\hline
\endhead
\hline \multicolumn{6}{r}{{Continued on next page}}\\
\endfoot
\hline
\endlastfoot
American Indian or Alaska Native & \highlightFail{Fail} & 1.1 & 0.00 & \highlightKeep{-0.15} & \highlightFail{0.00} \\ 
\hline
Anemia diagnosis diagnosed & \highlightFail{Fail} & 17.8 & 0.00 & \highlightFail{-0.66} & \highlightFail{0.00} \\ 
\hline
Asian & \highlightFail{Fail} & 6.2 & 0.00 & \highlightFail{-0.36} & \highlightFail{0.00} \\ 
\hline
Black or African American & \highlightFail{Fail} & 13.6 & 19.09 & \highlightKeep{0.15} & \highlightFail{1.40} \\ 
\hline
Black or African American diagnosed & \highlightPass{Pass} & 18.0 & 16.39 & \highlightKeep{-0.04} & \highlightKeep{0.91} \\ 
\hline
Chronic kidney disease diagnosis diagnosed & \highlightNoRef{No Reference} &  &  &  &  \\ 
\hline
Coronary artery disease diagnosis diagnosed & \highlightNoRef{No Reference} &  &  &  &  \\ 
\hline
Depression diagnosis diagnosed & \highlightFail{Fail} & 19.0 & 0.00 & \highlightFail{-0.68} & \highlightFail{0.00} \\ 
\hline
Fatty liver disease diagnosis diagnosed & \highlightNoRef{No Reference} &  &  &  &  \\ 
\hline
Female & \highlightFail{Fail} & 50.8 & 59.85 & \highlightKeep{0.18} & \highlightFail{1.18} \\ 
\hline
Hispanic or Latino & \highlightPass{Pass} & 18.9 & 18.02 & \highlightKeep{-0.02} & \highlightKeep{0.95} \\ 
\hline
Hispanic or Latino diagnosed & \highlightFail{Fail} & 17.0 & 14.38 & \highlightKeep{-0.07} & \highlightFail{0.85} \\ 
\hline
Hyperlipidemia diagnosis diagnosed & \highlightPass{Pass} & 70.0 & 74.32 & \highlightKeep{0.10} & \highlightKeep{1.06} \\ 
\hline
Hypertension diagnosis diagnosed & \highlightFail{Fail} & 70.0 & 0.00 & \highlightFail{-2.16} & \highlightFail{0.00} \\ 
\hline
Male & \highlightFail{Fail} & 49.5 & 37.08 & \highlightFail{-0.25} & \highlightFail{0.75} \\ 
\hline
Male diagnosed & \highlightFail{Fail} & 51.0 & 13.40 & \highlightFail{-0.88} & \highlightFail{0.26} \\ 
\hline
Native Hawaiian or Other Pacific Islander & \highlightFail{Fail} & 0.2 & 0.00 & \highlightKeep{-0.06} & \highlightFail{0.00} \\ 
\hline
Not Hispanic or Latino & \highlightPass{Pass} & 81.3 & 78.07 & \highlightKeep{-0.08} & \highlightKeep{0.96} \\ 
\hline
Not Hispanic or Latino diagnosed & \highlightFail{Fail} & 66.0 & 11.17 & \highlightFail{-1.36} & \highlightFail{0.17} \\ 
\hline
Osteoarthritis diagnosis diagnosed & \highlightFail{Fail} & 29.5 & 0.00 & \highlightFail{-0.91} & \highlightFail{0.00} \\ 
\hline
Other & \highlightFail{Fail} & 6.2 & 0.00 & \highlightFail{-0.36} & \highlightFail{0.00} \\ 
\hline
Peripheral vascular disease diagnosis diagnosed & \highlightNoRef{No Reference} &  &  &  &  \\ 
\hline
Retinopathy diagnosis diagnosed & \highlightFail{Fail} & 28.5 & 0.50 & \highlightFail{-0.87} & \highlightFail{0.02} \\ 
\hline
Sleep apnea diagnosis diagnosed & \highlightFail{Fail} & 18.0 & 0.00 & \highlightFail{-0.66} & \highlightFail{0.00} \\ 
\hline
Stroke diagnosis diagnosed & \highlightFail{Fail} & 20.0 & 0.00 & \highlightFail{-0.71} & \highlightFail{0.00} \\ 
\hline
White & \highlightFail{Fail} & 75.8 & 55.22 & \highlightFail{-0.44} & \highlightFail{0.73} \\ 
\hline
White diagnosed & \highlightFail{Fail} & 58.0 & 9.69 & \highlightFail{-1.19} & \highlightFail{0.17} \\ 
\hline
check incidence & \highlightFail{Fail} & 1.4 & 5.86 & \highlightFail{0.24} & \highlightFail{4.19} \\ 
\hline
check prevalence & \highlightPass{Pass} & 11.3 & 11.80 & \highlightKeep{0.02} & \highlightKeep{1.04} \\ 
\hline
check data types & \highlightFail{Fail} &  &  &  &  \\ 
\hline
\end{longtable}

\section{Drug Tests}

\begin{longtable}{p{7.5cm} c c c c c}
\label{tab:drugs}\\
\textbf{Test name} & \textbf{Result} & \textbf{Expected value} & \textbf{Actual value} & \textbf{SMD} & \textbf{ratio}\\
\hline
\endfirsthead
\multicolumn{6}{c}{{\bfseries Continued from previous page}}\\
\hline
\textbf{Test name} & \textbf{Result} & \textbf{Expected value} & \textbf{Actual value} & \textbf{SMD} & \textbf{ratio}\\
\hline
\endhead
\hline \multicolumn{6}{r}{{Continued on next page}}\\
\endfoot
\hline
\endlastfoot
A foreign key to a standard concept identifier for the drug concept 1 & \highlightFail{Fail} & 48.6 & 0.00 & \highlightFail{-1.38} & \highlightFail{0.00} \\ 
\hline
A foreign key to a standard concept identifier for the drug concept 2 & \highlightNoRef{No Reference} &  &  &  &  \\ 
\hline
A foreign key to a standard concept identifier for the drug concept 3 & \highlightPass{Pass} & 48.6 & 0.00 &  &  \\ 
\hline
check data types & \highlightFail{Fail} &  &  &  &  \\ 
\hline
\end{longtable}

\section{Measurement Tests}

\begin{longtable}{p{7.5cm} c c c c c}
\label{tab:measurements}\\
\textbf{Test name} & \textbf{Result} & \textbf{Expected value} & \textbf{Actual value} & \textbf{SMD} & \textbf{ratio}\\
\hline
\endfirsthead
\multicolumn{6}{c}{{\bfseries Continued from previous page}}\\
\hline
\textbf{Test name} & \textbf{Result} & \textbf{Expected value} & \textbf{Actual value} & \textbf{SMD} & \textbf{ratio}\\
\hline
\endhead
\hline \multicolumn{6}{r}{{Continued on next page}}\\
\endfoot
\hline
\endlastfoot
Blood urea nitrogen measurement & \highlightFail{Fail} & 95\% & 0.00 &  &  \\ 
\hline
Hemoglobin A1c measurement & \highlightFail{Fail} & 95\% & 60.64 &  &  \\ 
\hline
Lipid panel & \highlightFail{Fail} & 95\% & 0.00 &  &  \\ 
\hline
Serum alanine aminotransferase measurement & \highlightFail{Fail} & 95\% & 0.00 &  &  \\ 
\hline
Serum albumin measurement & \highlightFail{Fail} & 95\% & 0.00 &  &  \\ 
\hline
Serum alkaline phosphatase measurement & \highlightFail{Fail} & 95\% & 0.00 &  &  \\ 
\hline
Serum aspartate aminotransferase measurement & \highlightFail{Fail} & 95\% & 0.00 &  &  \\ 
\hline
Serum aspartate aminotransferase measurement 1 & \highlightFail{Fail} & 95\% & 0.00 &  &  \\ 
\hline
Serum bicarbonate measurement & \highlightFail{Fail} & 95\% & 0.00 &  &  \\ 
\hline
Serum bilirubin measurement & \highlightFail{Fail} & 95\% & 0.00 &  &  \\ 
\hline
Serum calcium measurement & \highlightFail{Fail} & 95\% & 0.00 &  &  \\ 
\hline
Serum chloride measurement & \highlightFail{Fail} & 95\% & 10.12 &  &  \\ 
\hline
Serum cholesterol measurement & \highlightFail{Fail} & 95\% & 7.86 &  &  \\ 
\hline
Serum creatinine measurement & \highlightFail{Fail} & 95\% & 0.00 &  &  \\ 
\hline
Serum creatinine measurement 1 & \highlightFail{Fail} & 95\% & 0.00 &  &  \\ 
\hline
Serum glucose measurement & \highlightFail{Fail} & 95\% & 85.25 &  &  \\ 
\hline
Serum high density lipoprotein measurement & \highlightFail{Fail} & 95\% & 48.28 &  &  \\ 
\hline
Serum lactate dehydrogenase measurement & \highlightFail{Fail} & 95\% & 0.00 &  &  \\ 
\hline
Serum low density lipoprotein measurement & \highlightFail{Fail} & 95\% & 52.84 &  &  \\ 
\hline
Serum magnesium measurement & \highlightFail{Fail} & 95\% & 0.00 &  &  \\ 
\hline
Serum phosphate measurement & \highlightFail{Fail} & 95\% & 0.00 &  &  \\ 
\hline
Serum potassium measurement & \highlightFail{Fail} & 95\% & 0.00 &  &  \\ 
\hline
Serum potassium measurement 1 & \highlightFail{Fail} & 95\% & 0.00 &  &  \\ 
\hline
Serum sodium measurement & \highlightFail{Fail} & 95\% & 0.00 &  &  \\ 
\hline
Serum triglycerides measurement & \highlightFail{Fail} & 95\% & 0.00 &  &  \\ 
\hline
Serum uric acid measurement & \highlightFail{Fail} & 95\% & 0.00 &  &  \\ 
\hline
Serum uric acid measurement 1 & \highlightFail{Fail} & 95\% & 0.00 &  &  \\ 
\hline
Urine microalbumin measurement & \highlightFail{Fail} & 30.0 & 74.61 &  &  \\ 
\hline
Urine microalbumin measurement 1 & \highlightFail{Fail} & 95\% & 74.66 &  &  \\ 
\hline
check data types & \highlightFail{Fail} &  &  &  &  \\ 
\hline
\end{longtable}
\newpage

\title{\textbf{Supplementary material}}
\author{}
\date{}

\maketitle

\tableofcontents
\listoftables    
\listoffigures

\renewcommand{\tablename}{Supplementary Table}
\renewcommand{\figurename}{Supplementary Figure}

\renewcommand{\thetable}{S\arabic{table}}
\renewcommand{\thefigure}{S\arabic{figure}}
\renewcommand{\thesection}{S\arabic{section}}

\newpage
\section{Prompt engineering}
\label{supsec:prompt_eng}
Prompts are used to extract information from LLMs. We structured the prompts to be as detailed as possible for the LLM to perform the required task. While it is common to keep context history in API calls, i.e. sending all previous messages and replies, we found that keeping only the system prompt and the required task yields better results, which aligns with the findings of Mirzadeh et al.\cite{mirzadeh_gsm-symbolic_2024} and Narayan et al.\cite{narayan_can_2022}. The used prompts are available in Supplementary Tables~\ref{tab:test_suggestion_prompts}, \ref{tab:test_gen_prompts} and \ref{tab:test_val_prompts}.
\\

\section{User provided information and unit test example}
\label{supsec:specs}
\begin{figure}
    \centering
    \includegraphics[width=0.45\linewidth]{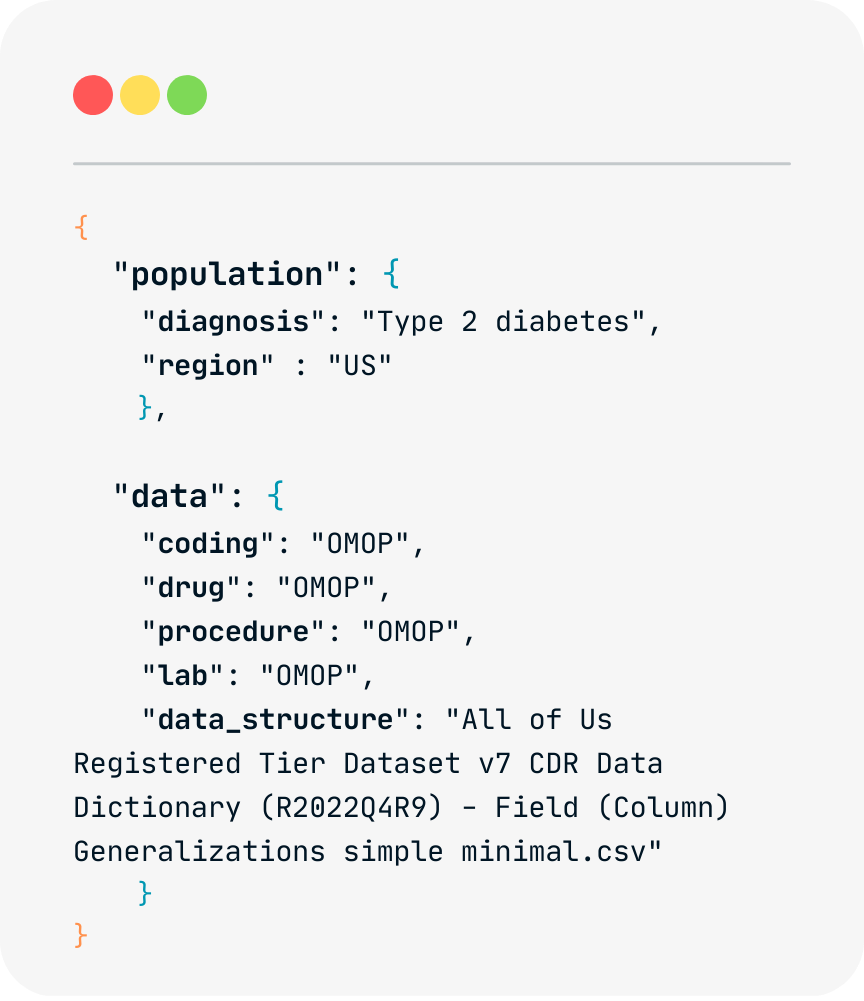}
    \caption{\textbf{An example of user provided study parameters in JSON format}, illustrating the defined cohort, the coding system, and the reference path to the data schema (All of Us).}
    \label{fig:json_def_example}. 
\end{figure}

\begin{figure}
    \centering
    \includegraphics[width=1\linewidth]{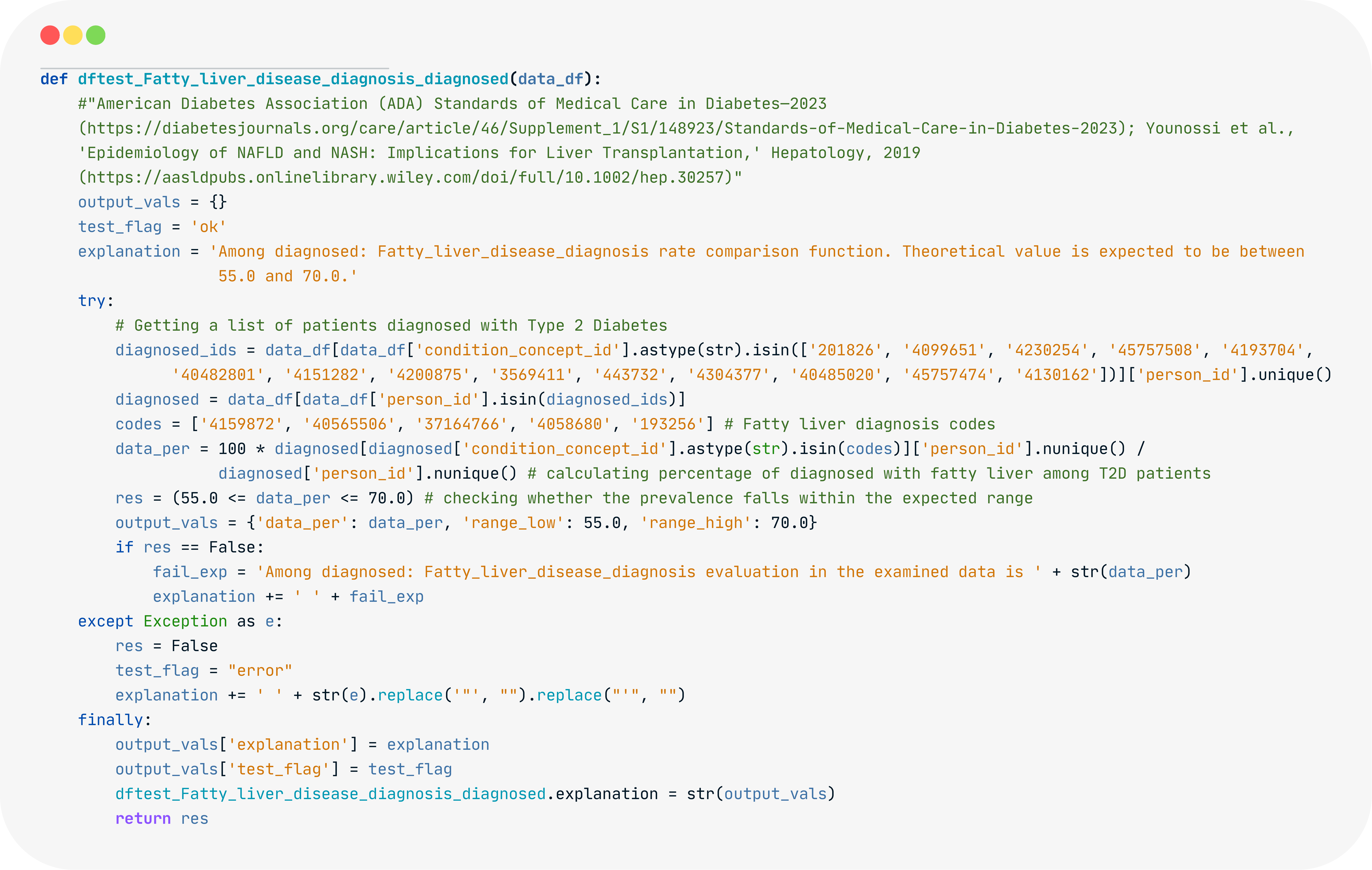}
    \caption{\textbf{Example unit test for validating expected data characteristics.}}
    \label{fig:output_test}
\end{figure}

Figure \ref{fig:json_def_example} illustrates the data provided by the user. This information is used by \mdpt{} to retrieve contextually relevant statistics. Our implementation uses an LLM (GPT-4o, model version 2024-08-06, with a temperature of 0 and seed set to 37, accessed via the Azure OpenAI API service, API version 2024-05-01-preview).
\mdpt{} uses these definitions to generate unit tests. An example of such unit test can be found in \ref{fig:output_test}.

\subsubsection*{Test quality validation}
\label{sec:test_quality_val}
To assess test quality, we evaluated reference validity and grounding accuracy.
Grounding accuracy was based on source trustworthiness and correctness of the cited reference values. A source was deemed invalid if it led to a nonexistent page or was unrelated to the studied condition. Reference values were categorized as follows:

\begin{itemize}
    \item \textbf{Accurate:} Matches the expected value from a credible source.
    \item \textbf{Inaccurate:} Within 2 standard deviations (SD) of the expected value, or a point estimate where a range is expected.
    \item \textbf{Incorrect:} Deviates by more than 2 SD from the expected value.
    \item \textbf{Qualitative:} Described as a trend or strength of association rather than a numeric value.
    \item \textbf{Irrelevant:} No relevant reference found for the required statistic.
\end{itemize}

Categories were grouped into "correct" and "non-correct" based on the accuracy of the reference values: the "correct" category includes the "accurate" group, while "non-correct" includes all other groups.
Test outcomes were considered passed if they matched the reference, and failed otherwise. Test quality results are presented in Table \ref{tab:test_quality}.

\begin{table}
\small
    \begin{tabular}{lcccccccc}
    \hline
        & 
         & \textbf{Valid References}
          & \textbf{Total Tests}
        & \textbf{Accurate} 
        & \textbf{Inaccurate}
        & \textbf{Incorrect}
        & \textbf{Qualitative} 
        & \textbf{Irrelevant} \\

         \hline
      & \textbf{\ttd{}}  & 88.1\% &  60 &  59.3\% & 16.9\% & 20.3\% & 0\% & 3.9\%  \\
                    \hline
      & \textbf{\ckd{}}  & 43.9\% &  67 & 60.6\%  & 18.2\%  & 12.2\%  & 3.0\%  & 6.1\% \\
                    \hline
       & \textbf{\htn{}}  & 93.0\% & 73  & 59.7\%  & 6.9\% &  26.4\%  & 4.2\%  &  2.8\%  \\
                     \hline
      & \textbf{\chf{}} & 35.2\% &  55  & 42.6\%  &  22.2\%  &  20.4\% &  1.9\%  &  13.0\%  \\
                        \hline
    \end{tabular}
    \caption{\textbf{Reference Validity and Grounding Accuracy of \mdpt{}-Generated Tests Across Four Cohorts.} Summary of test quality evaluation for each condition-specific test suite, reporting the percentage of tests with valid references and the distribution of grounding accuracy categories: accurate, inaccurate (within 2 SD), incorrect (greater than 2 SD), qualitative (non-numeric), and irrelevant (no matching reference). Results reflect the variability in literature coverage and grounding success across datasets and clinical contexts. For each table entry other than the overall test count, we expressed the number of tests in the specified category as a proportion of all tests, excluding the metadata test.}
    \label{tab:test_quality}
\end{table}

\newpage

\begin{longtable}{p{15cm}}
\caption{Test suggestion prompts}\\
\label{tab:test_suggestion_prompts}\\
\hline
        \textbf{Prior Bing searches}\\
    \hline
        \begin{verbatim}
searchs = query_bing(f'prevalence of {diagnosis} in {region}')
searchs1 = query_bing(f'incidence of {diagnosis} in {region}')
searchs2 = query_bing(f'mortality rate of {diagnosis} in {region}')
searchs3 = query_bing(f'lifetime risk of {diagnosis} in {region}')
searchs4 = query_bing(f'survival rate of {diagnosis} in {region}')
search1 = query_bing(f'age distribution among {diagnosis} patients in {region}')
search2 = query_bing(f'mean age and standard deviation of {diagnosis} in {region}')
search3 = query_bing(f'Total {region} population size')
search4 = query_bing(f'Number of diagnosed with {diagnosis} in {region}')
        \end{verbatim}\\
    \hline
       \textbf{System prompt} \\
    \hline
       \begin{verbatim}
You are a public health and medical expert. Answer as concisely as possible. 
Answer for the provided diagnosis, region, coding system, and drug classification
system. Make sure the coding in all your answers fits the requirements. The 
following questions are about {diagnosis} in {region}. When searching for 
references, have them as up-to-date as possible. Search for regional data and 
provide references. If regional data is unavailable, reply with global data. If 
there is no available data, answer with 'Data is unavailable'. If you are unsure, 
answer with 'I am unsure.'
Work step by step. 
**When asked for diagnosis codes, provide only the rerquired codes and don't 
\end{verbatim}\\
\begin{verbatim}
randomly assign codes.**
**When asked for any classification codes, provide the required codes without 
truncation.**
**Stick to the required coding systems.**
If asked about diagnosis/drug/procedure codes, the fields in the reference data are 
'concept_id' which includes OMOP codes, 'concept_name' which are the names of the 
codes, 'concept_code' which are the source vocabulary codes, 'vocabulary_id' is the
name of the source vocabulary.
In OMOP coding, the standard dictionary is SNOMED and RxNorm, and the required field
is concept_id. For exery other coding, the required field is concept_code.
For demographic variables, always include the race, gender and ethnicity 
vocabularies.
When asked for csv tables, divide them using a vertical line |. Add the required 
title before the csv, exactly as written in the request. 
Please avoid other symbols arount the title. Give me your best answer and provide 
references. When asked for Python code, CSVs or text, always format them using 
triple backticks as follows: '```python', '```csv' and '```text', respectively, and
end with '```'. At the top of each block (Python code, CSV, or text), include the 
required title or filename as part of the response. This title or filename must 
appear at the start of each block:

```python
Filename: example file name
# your python code here
```

or 

```csv
Title: example title
# your csv code here
```

If no filename or title was provided in the reqest, please avoid including a 
filename in the response.
Please provide references including a **valid** url or a publication for each answer.
Please confirm understanding of these instructions without replying. 
    \end{verbatim}
        \\
       \hline
       \textbf{Prompts - regional statistics} \\
       \hline
       \begin{verbatim}
What is the prevalence of {diagnosis} in {region}? Provide the percentage of 
diagnosed. 
Prioritize more recent data, preferably from government sources.
Please use the following information and provide numbers:

{searchs['webPages']['value']}

Numbers are mandatory. Please provide references.
       \end{verbatim} \\
       \hline
        \begin{verbatim}
What is the incidence of {diagnosis} in {region}? Provide the percentage of 
diagnosied. 
Prioritize more recent data, preferably from government sources.
Please use the following information and provide numbers:

{searchs1['webPages']['value']}

Numbers are mandatory. Please provide references.
        \end{verbatim}\\
        \hline
        \begin{verbatim}
What is the mortality rate of people with {diagnosis} in {region}? Provide the 
percentage. 
Prioritize more recent data, preferably from government sources.
Please use the following information and provide numbers:

{searchs2['webPages']['value']}

Numbers are mandatory. Please provide references.
        \end{verbatim}\\
        \hline
        \begin{verbatim}
What is the lifetime risk of developing {diagnosis} in {region}? Provide the 
percentage. 
Prioritize more recent data, preferably from government sources.
Please use the following information and provide numbers:

{searchs3['webPages']['value']}

Numbers are mandatory. Please provide references.
        \end{verbatim}\\
        \hline
        \begin{verbatim}
What is the survival rate of people with {diagnosis} in {region}? Provide the 
percentage. 
Prioritize more recent data, preferably from government sources.
Please use the following information and provide numbers:

{searchs4['webPages']['value']}

Numbers are mandatory. Please provide references.

        \end{verbatim}\\
        \hline
    \begin{verbatim}
What is the total population size in {region}? Provide the number in millions. 
Prioritize more recent data, preferably from government sources.
Please use the following information and provide numbers:

{search3['webPages']['value']}

Numbers are mandatory. Please provide references.
    \end{verbatim}\\
    \hline
    \begin{verbatim}
Please provide statistics for people diagnosed with {diagnosis} in {region}. Include
Prevalence, Incidence, Mortality rate, Lifetime risk, and Survival. 
Provide all statistics in percentage. Please use your previous replies as reference.

Arrange them in a csv table with a vertical line as delimiter. Please add 
references. Add 'Title: statistics_in_{region}' before the CSV. 
Do not add any other symbols (such as dashes or hypens) to the table. 
Structure the table as follows: Prevalence | Incidence | Mortality rate | 
Lifetime risk | Survival | total population size in millions | reference.
Give me your best answer. Please provide references.
    \end{verbatim}\\
    \hline
   \begin{verbatim}
Please provide mean and standard deviation of {diagnosis} diagnosis age in {region}. 
Please use the following information and prioritize more recent data:

{search1['webPages']['value']}

{search2['webPages']['value']}

If there is no information in the provided data, please use your knowledge.
Numbers are mandatory. Please provide references.
   \end{verbatim}\\
   \hline
    \begin{verbatim}
Please provide the number of diagnosed with {diagnosis} in {region}. please write 
a full number (e.g instead of '10 million' write 10000000). Use the following 
reference:

{search4['webPages']['value']}

If there is no information in the provided data, please provide it based on the 
prevalence and total population size.

Numbers are mandatory. Please provide references.
    \end{verbatim}\\
    \hline
    \begin{verbatim}
Please provide the age statistics for {diagnosis} in {region}. Include mean age, 
standard deviation, and count.
Use your previous replies as reference. If there is no data in the provided 
information, please use your knowledge.           

Arrange them in a csv table with a vertical line as delimiter. 
Please add references. Add 'Title: age_stats_in_{region}' before the CSV. Do not add
any other symbols (such as dashes or hypens) to the table. 
Give me your best answer. Structure the table as follows: mean_age | std | count |
reference. Make sure all columns are filled. Please provide references for the 
opercentage values.
    \end{verbatim}\\
    \hline
    \textbf{List of Comorbidities}\\
    \hline
    \begin{verbatim}
Please provide a list of 15 most prevalent comorbidities among {diagnosis} patients. 
The concept names of the comorbidities should appear as they appear in {coding} 
dignosis codes.
 provide a comma separated list of the commorbidities separated by '```list' and 
 '```'.
    \end{verbatim}\\
    \hline
    \textbf{Comorbidities - vector database and Bing queries}\\
    \hline
    \begin{verbatim}
omorbidities.append(json_to_df(get_concepts('concept_name: ' + comorbidity + '
domain_id: Condition ' + 'vocabulary_id: ' + coding + ' standard_concept: S', 
n_results=100)))
    \end{verbatim}\\
    \hline
    \begin{verbatim}
search = query_bing(f'Prevalence of {comorbidity} among {diagnosis} patients in
{region}')
    \end{verbatim}\\
    \hline
    \textbf{Comorbidities - prompts}\\
    \hline
    \begin{verbatim}
Please provide a comprehensive and full list of {coding} codes for {comorbidity}.
Provide a full list of codes without truncation.
Use the following table:

{comorbidities[idx]}

Please include all possible options. Start with general {comorbidity} {coding} 
codes, preferably standard ones (with "S"). 
The csv table should include "{coding} codes" and "comorbidity" columns. Write 
comorbidity name and a list of all relevant {coding} codes in one row. Separate the 
csv using a vertical line as delimiter.
First include all possible {comorbidity} {coding} codes without complications. Work 
step by step. Do not generate random codes.Use only the ones provided. Give me your
best answer.
    \end{verbatim}\\
    \hline
    \begin{verbatim}
Please add a "prevalence" column with prevalence of {comorbidity} among {diagnosis} 
patients. 
Add a reference column with reference to the prevalence value. Prioritize more 
recent data, preferably from government sources.
use the following information:

{search['webPages']['value']}

Give me your best answer. Do not change or remove any other column. Separate the csv
using a vertical line as delimiter.
    \end{verbatim}\\
    \hline
    \begin{verbatim}
Please provide a full table of comorbidities, based on the following table:
                      
{comorbidities1}


provide a table with the following columns: comorbidity|{coding} codes|prevalence.
Wrap the csv in '```csv' and '```'.
Use a vertical line as delimiter. Add 'Title: comorbidities_in_{region}' before the
CSV.
    \end{verbatim}\\
    \hline
    \textbf{List of drugs}\\
    \hline
    \begin{verbatim}
Please provide a list of 15 specific drug names and/or ingredients used to treat 
{diagnosis}. 
Be as specific as possible. The concept names of the drugs should appear as they 
appear in {drug} drug codes.
Provide a comma separated list of the above drugs separated by '```list' and '```'.
    \end{verbatim}\\
    \hline
    \textbf{Drugs - vector database and Bing queries}\\
    \hline{}
    \begin{verbatim}
drugs.append(json_to_df(get_concepts('concept_name: ' + drug_name + 'domain_id: 
Drug' + 'vocabulary_id: ' + drug + ' standard_concept: S', n_results=140)))

    \end{verbatim}\\
\hline
\begin{verbatim}
search = query_bing(f'Percentage of {diagnosis} patients prescribed {drug_name} in 
{region}')
\end{verbatim}\\
\hline
\textbf{Drugs - prompts}\\
\hline
\begin{verbatim}
Please provide a comprehensive list of {drug} codes for {drug_name}. Provide a full 
list of codes without truncation.
Use the following table:

{drugs[idx]}

Please include all possible options. Start with general {drug_name} {drug} codes, 
preferably standard ones (with "S"). 
The csv table should include "{drug} codes" and "drug" columns. Write a list of all
relevant {drug_name} {drug} codes under "{drug} codes", and drug name under "drug",
in one row. Work step by step. Do not generate random codes, use only the ones 
provided. 
Arrange them in a csv table with a vertical line as delimiter. Give me your best 
answer.
\end{verbatim}\\
\hline
\begin{verbatim}
Please add a "percentage" column with percentage of {diagnosis} patients treated 
with the drug and a "reference" column for the percentage information source with 
the relevant references. Prioritize more recent data, preferably from government 
sources.
use the following information:

{search['webPages']['value']}
  
Give me your best answer. Do not change or remove any other column.
\end{verbatim}\\
\hline
\begin{verbatim}
Please provide a full table of drugs, based on the following table:

{drugs1} 

Please create a csv with the following columns: {drug} codes|drug|percentage|reference. 
Wrap the csv in '```csv' and '```'.
Please arange it so all similar codes are aggregated under the '{drug} code' column
without truncation, description of the same drug name under 'drug'.
Use a vertical line as delimiter. Add 'Title: drug_codes_in_{region}' before the CSV.
\end{verbatim}\\
\hline
\textbf{List of labs and procedures}\\
\hline
\begin{verbatim}
Please provide a list of top 25 specific lab tests and procedure names performed on
{diagnosis} patients  in {region}. Be as specific as possible with procedure or lab
test name. Avoid panel names, include individual tests. 
The concept names of the procedures and lab tests should appear as they appear in 
{procedure} procedure codes and {lab} terminology.
Provide a comma separated list of the lab tests and procedures separated by 
'```list' and '```'.
\end{verbatim}\\
\hline
\textbf{Labs and Procedures - vector database and Bing quieries}\\
\hline
\begin{verbatim}
json_to_df(get_concepts('concept_name: ' + lab + 'domain_id: measurement' + 
'vocabulary_id: ' + lab + ' standard_concept: S', n_results=70))
\end{verbatim}\\
\hline
\begin{verbatim}
json_to_df(get_concepts('concept_name: ' + lab + 'domain_id: procedure' + 
'vocabulary_id: ' + procedure + ' standard_concept: S', n_results=70))
\end{verbatim}\\
\hline
\begin{verbatim}
search = query_bing(f'Normal range of {lab} in {region}')
\end{verbatim}\\
\hline
\textbf{Labs and procedures - prompts}\\
\hline
\begin{verbatim}
Please provide a comprehensive list of {lab} codes for the following lab test or 
procedure: 

{Labs[idx]}

Arrange them in a csv table with a vertical line as delimiter. Please include all
possible options.
The csv table should include "procedure codes" and "procedure" columns. Write a list
of all relevant {lab} or {procedure} codes under "procedure codes", and procedure 
name under "procedure", in one row.
\end{verbatim}\\
\begin{verbatim}
Work step by step. Do not generate random codes. Provide a full list of codes 
without truncation.
Use only the ones provided. Give me your best answer.
\end{verbatim}\\
\hline
\begin{verbatim}
Please add a "normal range" column with normal ranges of results for each of the 
lab tests or procedures. 
Prioritize more recent data, preferably from government sources. Please use the 
following:

{search['webPages']['value']}

Add a "reference" column for the normal range information source with the relevant
references.
Give me your best answer. Do not change or remove any other column.
\end{verbatim}\\
\hline
\begin{verbatim}
Please provide a full table of lab tests and procedures, based on the following 
table:

{Labs1} 

Please create a csv with the following columns: 
{coding} codes|procedure|normal range.
Wrap the csv in '```csv' and '```'. Use a vertical line as delimiter. 
Add 'Title: 'procedure_codes_in_{region}' before the CSV.
\end{verbatim}\\
\hline
\textbf{General demographic search}\\
\hline
\begin{verbatim}
search1 = query_bing(f'Percentage of {demo} in {region}')
\end{verbatim}\\
\hline
\begin{verbatim}
search2 = query_bing(f'Percentage of {demo} diagnosed with {diagnosis} in {region}')
\end{verbatim}\\
\hline
\textbf{Gender or race or ethnicity - vector database and Bing queries}\\
\hline
\begin{verbatim}
search = query_bing(f'Gender identity distribution in {region}')
\end{verbatim}\\
\hline
\begin{verbatim}
gender_list.append(json_to_df(get_concepts("concept_name: " + demo + 'domain_id:
Gender' + 'standard_concept: S', n_results=50)))

or

race_list.append(json_to_df(get_concepts("concept_name: " + demo + 'domain_id: Race'
+ 'standard_concept: S', n_results=70)))

or

ethnicity_list.append(json_to_df(get_concepts("concept_name: " + demo + 'domain_id:
Ethnicity' + 'standard_concept: S', n_results=50)))

\end{verbatim}\\
\hline
\textbf{List of gender values}\\
\hline
\begin{verbatim}
List all possible {coding} concept names for all possible gender options in 
{region}. 
Prioritize more recent data, preferably from government sources. Please use the 
following as reference as well:

{search['webPages']['value']}

Provide a comma separated list of the concepts separated by '```list' and '```'.
Just write it as single list, do not declare the category title (gender, race, 
ethnicity).
\end{verbatim}\\
\hline
\textbf{Gender - prompts}\\
\hline
\begin{verbatim}
Please provide a comprehensive list of {gender or race or ethnicity} {coding} codes
according to:
   {gender_list[idx] or race_list[idx] or ethnicity_list[idx]}.

Arrange them in a csv table with a vertical line as delimiter.
Please include all possible options for {demo}. Drop irrelevant options.
The csv table should include "{coding} code" and "demographic feature" columns. 
Each categorical feature - {gender or race or ethnicity} value -  should appear in 
a separate row.
For each row, write all the relevant {coding} codes in the "{coding} code" column 
without truncation, and the appropriate demographic feature in "demographic feature"
column. 
"demographic feature" column should contain the specific gender, race or ethnicity 
subtype concept name.
Work step by step. Do not generate random codes. Use only the ones provided. 
Give me your best answer.
\end{verbatim}\\
\hline
\begin{verbatim}
Please add "percentage" column with the percentage of {demo} in {region}. 
Prioritize more recent data, preferably from government sources. Please use the
following:
  
{search1['webPages']['value']}

Add a reference in the "reference" column for the percentage information source 
with the relevant references.
Give me your best answer. Do not change or remove any other column.
\end{verbatim}\\
\hline
\begin{verbatim}
Please add "diagnosed percentage" column with the percentage of {demo} diagnosed 
with {diagnosis} in {region}. 
Prioritize more recent data, preferably from government sources. Please use the 
following:

{search2['webPages']['value']}
\end{verbatim}\\
\begin{verbatim}
Add a reference in the "reference" column for the diagnosed percentage information
source with the relevant references.
Give me your best answer. Do not change or remove any other column
\end{verbatim}\\
\hline
\textbf{List of race values}\\
\hline
\begin{verbatim}
List all possible {coding} concept names for all possible race options in {region}.
Provide a comma separated list of the concepts separated by '```list' and '```'.
Just write it as single list, do not declare the category title (gender, race, 
ethnicity).
\end{verbatim}\\
\hline
\textbf{List of ethnicity values}\\
\hline
\begin{verbatim}
List all possible {coding} concept names for all possible ethnicity options in 
{region}. 
Provide a comma separated list of the concepts separated by '```list' and '```'.
Just write it as single list, do not declare the category title (gender, race, 
ethnicity).
\end{verbatim}\\
\hline
\textbf{Stacking all demographics - Bing query}\\
\hline
\begin{verbatim}
search = query_bing(f'gender statistics in {region}')
\end{verbatim}\\
\hline
\begin{verbatim}
search1 = query_bing(f'race statistics in {region}')
\end{verbatim}\\
\hline
\begin{verbatim}
search2 = query_bing(f'ethnicity statistics in {region}')
\end{verbatim}\\
\hline
\textbf{Stacking all demographics - prompt}\\
\hline
\begin{verbatim}
Take the following table:

{demo_list1}
\end{verbatim}\\
\begin{verbatim}
Arrange it so that all similar codes are aggregated into a single column as a list 

under '{coding} code', description of the similar codes under 'demographic feature'.

Add population percentage in {region} under 'Percentage' column based on known 
statistics. For the statistics, provide **reliable and valid** references - a url 
or a publication.

use the following as reference:

{search['webPages']['value']}

{search1['webPages']['value']}

{search2['webPages']['value']}

Use a vertical line as delimiter.
**Do not** change or remove any other column. 
Add 'Title: demography_coding' before the CSV. 
Wrap the csv in '```csv' and '```'.
Give me your best answer. Provide references.
\end{verbatim}\\
\hline
\textbf{Stacking all demographics among diagnosed - Bing queries}\\
\hline
\begin{verbatim}
search = query_bing(f'gender statistics among patients with {diagnosis} in {region}')
\end{verbatim}\\
\hline
\begin{verbatim}
search1 = query_bing(f'race statistics among patients with {diagnosis} in {region}')
\end{verbatim}\\
\hline
\begin{verbatim}
search2 = query_bing(f'ethnicity statistics among patients with {diagnosis} in 
{region}')
\end{verbatim}\\
\hline
\textbf{Stacking all demographics among diagnosed - prompt}\\
\hline
\begin{verbatim}
Take the following table:

{demo_list1}

Arrange it so that all similar codes are aggregated into a single column as a list 
under '{coding} code', description of the similar codes under 'demographic feature'.
Add diagnosed population percentage in {region} under 'Percentage' column based on
known statistics. For the statistics, provide **reliable and valid** references - a 
url or a publication.
Make sure the statistics you provide are the percentage of diagnosed patients among
specific demographic groups.

use the following as reference:

{search['webPages']['value']}

{search1['webPages']['value']}

{search2['webPages']['value']}

Use a vertical line as delimiter.
**Do not** change or remove any other column. 
Add 'Title: diagnosed_demography_coding' before the CSV. 
Wrap the csv in '```csv' and '```'.
Give me your best answer. Provide references.
\end{verbatim}\\
\hline
\end{longtable}

\newpage

\begin{longtable}{p{15cm}}
\caption{Test generation prompts}\\
\label{tab:test_gen_prompts}\\
    \hline
       \textbf{System prompt} \\
    \hline
       \begin{verbatim}
       
We will create a table of values to evaluate. Please stick to the provided 
information, unless specifically asked to provide additional information which was 
not mentioned in the files. Include medical, drug, and procedure classification 
coding where relevant. Use {coding} for diagnosis classification codes, {drug} for 
drug classification codes, {procedure} for procedures, {lab} for lab measurement 
classification codes and {coding} and gender, race and ethnicity classification for
demographic coding. 
If unsure, respond with "I am unsure."
Make sure the CSVs do not contain anything that will prevent them from being loaded
into a pandas dataframe.

When asked for CSV tables:
Use | (vertical bar) as the separator.
Enclose the table in '```csv' and '```'.
Exclude any symbols other than those specified.
Separate the text and table clearly.
Create all tables with the following columns: 'flag','column name','codes','data 
type','value column name','description','expected value','expected value 
max','expected value min','range low','range high','comparison type','references'

If there are multiple classification codes for a specific concept, provide them as
a list in the 'codes' column (e.g, if there are multiple codes for a specific 
diagnosis, drug, procedure, lab test or a demographic property). 
\end{verbatim}\\
\begin{verbatim}
**Please create a separate row for each subpopulation (e.g. for gender - separate 
rows for male and for female, as well as for the different race and ethnicity 
options).**

Gender, race and ethnicity categories are in the data dictionary 'description' 
column and should be used as the possible code values for the demographic 
subpopulations.

If there is a single expected value, fill "expected value min" and "expected value 
max" 'with np.nan', and fill only "expected value" with the value.
"expected value min" is used when the expected result is greater than the expected 
value, "expected value max" is used when the expected result is less than the 
expected value.
If we expect a range of values, fill "range low" and "range high".
'column name' should contain the name of the column from the provided data dictionary 
that holds the classifiction codes.
"value column name" should contain the name of the column from the provided data 
dictionary that holds the recorded value for the test or procedure or measurement, 
if applicable
If there are multiple medical, drug, lab or procedure classification codes related 
to a specific concept name, add them as a list in the relevant classification code-
related column. 
Ensure all tables can be concatenated.

For date-related data, use columns with datetime values, if available.
For each diagnosis ({coding}), drug ({drug}), measurement/lab test ({lab}) and 
procedure ({procedure}) code, provide **only** the relevant meaning/concept name in
the 'description' column. Double check the meaning and provide references for the 
statistical values.
Provide your response step by step and offer your best possible answer. 
Provide references for all descriptions, code meanings, and for the statistical 
values.

When asked for CSVs or text, always format them using triple backticks as follows: 
'```csv' or '```text', respectively, and end with '```'. At the top of each block 
(CSV, or text), include the required title or filename as part of the response. 
This title or filename must appear at the start of each block:

```csv
Title: example title
# your csv code here
```

If no filename or title was provided in the reqest, please avoid including a 
filename in the response.
Please confirm understanding of these instructions without replying.
    \end{verbatim}
        \\
       \hline
       \textbf{Assistance table for correct column naming} \\
       \hline
       \begin{verbatim}
You are given a database schema where each row specifies a table name, column name,
and description.
Given the following schema:

{data_struct}

Identify the table name, column name, and description for the field related to 
patient's date of birth. Return the result in the following format:

Table: <table_name>
Column: <column_name>
Description: <description>
       \end{verbatim} \\
       \hline
        \begin{verbatim}
Given the following schema:

{data_struct}

Identify the table name, column name, and description for the field that stores 
**diagnosis code** or **concept identifier** entries. 
Focus on fields that reference standardized vocabularies or classification systems.
Return the result in the following format:

Table: <table_name>
Column: <column_name>
Description: <description>
        \end{verbatim}\\
        \hline
        \begin{verbatim}
Look at the demography and diagnoses using the relevant columns from:

{data_struct}

Identify the table name, column name, and description for the field related to 
"diagnosis first date". Return the result in the following format:

Table: <table_name>
Column: <column_name>
Description: <description>
        \end{verbatim}\\
        \hline
        \begin{verbatim}
Look at the demography and diagnoses using the relevant columns from:

{data_struct}

Identify the table name, column name, and description for the field related to 
"Person's unique identifier". Return the result in the following format:

Table: <table_name>
Column: <column_name>
Description: <description>
        \end{verbatim}\\
        \hline
        \begin{verbatim}
Based on all your previous answers, create a table with the following columns: 
'table_name', 'column_name', 'description', 'generic'. 
Populate the table with these specific descriptions: birth date, diagnosis code, 
diagnosis first date, and unique person identifier. 
Add the following to 'generic' column, according to the proper rows: 'personID', 
'birthDate', 'code', 'codeDate'. Inclue all of these.

The 'code' and 'codeDate' columns must refer to the {coding} medical coding system, 
excluding any other coding systems that may be present. 
'codeDate' should refer to the date and time of the diagnosis.
'birthDate' is the date of birth of the person.
All fields are mandatory. If a field is irrelevant, fill it with np.nan.
Wrap the table in a CSV code block using '```csv' and '```'. 
Add the title 'Title: convert_to_datastruct' right before the CSV block. 
Use a vertical bar '|' as the column separator, without any extra symbols in the 
table. 
Include all specified columns in the CSV, and prefer datetime columns where 
applicable for date-related data.
Work step by step. Make sure all column_name value`s are correct and appear in the 
data dictionary. Double check your reply.
Work step by step. Give me your best answer.
        \end{verbatim}\\
    \hline
    \textbf{Comorbidities}\\
    \hline
    \begin{verbatim}
We are going to check comorbidity prevalence among diagnosed. 
Please use the relevant tables and columns the following data dictinary:

{data_struct}

and the {coding} codes provided in the following reference file: 

{info} 
\end{verbatim}\\
\begin{verbatim}
For each comorbidity in the provided reference file, suggest expected values for 
patients diagnosed with {diagnosis} in {region}. 
Please use the description column in the data dictionary to describe the content of 
the column.
Each column/field name in the provided data dictionary can be evaluated as many times
as needed, for as many comorbidities as possible. 
Build a csv table with ALL of the following columns: 
'flag', 'column name', 'codes', 'data type','value column name','description',
'expected value','expected value max', expected value min', 'range low', 
'range high','comparison type','references'. 
 
All columns are mandatory. If a field is irrelevant, fill it with np.nan. Do not 
suggest random population statistics. Only use columns and tables from the provided 
data dictionary.
Use only codes in the reference file.
The 'flag' column should indicate the data type and should be filled with one of the
following values: 'demography', 'drug', 'procedure', 'lab' or 'diagnosis'.
'value column name' should contain the actual name of the column from the provided 
data dictionary that holds the value.
Before the CSV, include the title: "Title: expected_params_comorbidities". Do not 
separate it from the CSV code block.
Use the vertical bar | as the CSV separator.
Please include only relevant diagnosis codes in the 'codes' column.
Surround the CSV table with ```csv and ```.
In the 'column name' field, use the corresponding column name from the data 
dictionary where the comorbidity {coding} code can be found.
For the 'comparison type', specify whether the comparison is an 'exact value', 
'range', or 'percentage'. If it is a range, use numeric values only.
Do not suggest random population statistics. Only use columns and tables from the 
provided data dictionary.
In the 'codes' column, insert all the relevant {coding} medical codes from the 
reference file as a list. Each comorbidity should appear in a separate row. 
Include the code meaning in the 'description' column, double check the meaning and
provide references for the statistical values. 
Keep the 'description' column short, concise and specific with exact descripton of 
what is evaluated.
Expected values should be percentage of {diagnosis} patients with the comorbidity.
Use all of the provided information to complete the CSV. Include a full list of drug
codes.
Provide as many column/field name evaluations from the given data dictionary as 
possible.
Make sure the 'description' column includes the description of the column from the 
data dictionary, including the meaning of the corresponding {coding} codes. 
Provide references for the code meanings and for the statistical values. Work step by
step. Give me your best answer.
    \end{verbatim}\\
    \hline
    \textbf{Drugs}\\
    \hline
    \begin{verbatim}
We are going to evaluate drug use. Please use the relevant tables in the following 
data dictionary:

{data_struct}

and the {coding} codes provided in the following reference file:

{info}

For each column or field name in the provided data dictionary, suggest expected 
values for patients diagnosed with {diagnosis} in {region}. 
Please use the description column in the data dictionary to describe the content of 
the column.
Each field name can be used as many times as needed, for as many drugs as possible. 
Build a csv table with ALL of the following columns: 
'flag', 'column name', 'codes', 'data type','value column name',
'description','expected value','expected value max', 'expected value min', 
'range low', 'range high', 'comparison type','references'. 

All columns are mandatory. If a field is irrelevant, fill it with np.nan. Do not 
suggest random population statistics. Only use columns and tables from the provided 
data dictionary.
Use only codes in the reference file.

The 'flag' column should indicate the data type and should be filled with one of the
following values: 'demography', 'drug', 'procedure', 'lab' or 'diagnosis'.
'value column name' should contain the actual name of the column from the provided 
data dictionary that holds the value.
Before the CSV, include the title: "Title: expected_params_drugs". Do not separate it
from the CSV code block.
Use the vertical bar | as the CSV separator.
Surround the CSV table with ```csv and ```.
In the 'column name' field, use the corresponding column/field name from the data 
dictionary.
For the 'comparison type', specify whether the comparison is an 'exact value', 
'range', or 'percentage'. If it is a range, use numeric values only.
\end{verbatim}\\
\begin{verbatim}
Do not suggest random population statistics. Only use columns and tables from the 
provided data dictionary.
In the 'codes' column, insert all the relevant {drug} drug codes from the reference 
file as a list. Each drug should appear in a separate row. Include the code meaning 
in the 'description' column, double check the meaning and provide references for the
statistical values. 
Keep the 'description' column short, concise and specific with exact descripton of 
what is evaluated.
Expected values should be percentage of {diagnosis} patients who use the drug.
Use all of the provided information to complete the CSV.
Provide as many column/field evaluations from the given data dictionary as possible.
For date-related data, prefer columns with datetime values where applicable.
Make sure the 'description' column includes the description of the column from the 
data dictionary, including the meaning of the corresponding {coding} codes. 
Provide references for the code meanings and for the statistical values. Work step by
step. Give me your best answer.
    \end{verbatim}\\
    \hline
    \textbf{Procedures}\\
    \hline
    \begin{verbatim}
From the following data dictionary, please extract the table and column names which
contain lab test codes and recorded value:
    
    {data_struct}

Please include the table name, column name and description for the lab test code and
recorded value. 
Structure your answer as follows: 'Lab tests: table name, column name, description'.
    \end{verbatim}\\
    \hline
    \begin{verbatim}
From the following data dictionary, please extract the table and column names with 
procedure codes and recorded values in the provided data dictionary.

Please include the table name, column name and description for the lab test code and 
recorded value.
If there is no such column, please reply with 'no value column', and include only the
description and code columns.
Structure your answer as follows: 'Procedures: table name, column name, description'.
    \end{verbatim}\\
    \hline
    \begin{verbatim}
Based on your previous answers and the {procedure} procedure codes provided in the 
following reference file:

{info}

suggest expected values for the procedure in {region}.

Each column/field name can be used for as many procedures as needed.
Build a csv table with ALL of the following columns:
'flag', 'column name', 'codes', 'data type','value column name','description',
'expected value','expected value max', expected value min', 'range low', 
'range high','comparison type','references'. 

All columns are mandatory. If a field is irrelevant, fill it with np.nan. Do not 
suggest random population statistics. Only use columns and tables from the provided 
data dictionary.
Use only codes in the reference file.
The 'flag' column should indicate the data type and should be filled with one of the
following values: 'demography', 'drug', 'procedure', 'lab' or 'diagnosis'.
'value column name' should contain the field name in the provided data dictionary 
with the procedure value, if applicable. If it doesn't exist, fill it with 'np.nan'.
Please make sure the correct 'value column name' for procedure value is in the output 
csv, if applicable.
Before the CSV, include the title: "Title: expected_params_procedures". Do not 
separate it from the CSV code block.
\end{verbatim}\\
\begin{verbatim}
Use the vertical bar | as the CSV separator.
Surround the CSV table with ```csv and ```.
In the 'column name' field, use the corresponding column name from the data 
dictionary.
For the 'comparison type', specify whether the comparison is an 'exact value', 
'range', or 'percentage'. If it is a range, use numeric values only.
Do not suggest random numbers. Only use columns and tables from the provided data 
dictionary.
In the 'codes' column, insert all the relevant {procedure} procedure codes from the 
reference file as a list. 
Each procedure should appear in a separate row. 
Please include the code meaning in the 'description' column, double check the meaning
and provide references for the statistical values. 
Keep the 'description' column short, concise and specific with exact descripton of 
what is evaluated (e.g. 'urinary tract infection icd-9 codes').
In the values, include only the number, without percentage or units. Expected values
should be normal range of values for the evaluated procedure.
Use all of the provided information to complete the CSV.
Provide as many column evaluations from the given data dictionary as possible.
Make sure the 'description' column includes the description of the column from the 
data dictionary, including the meaning of the corresponding {coding} codes. 
Provide references for the code meanings and for the statistical values. Work step by
step. Give me your best answer.
    \end{verbatim}\\
    \hline
    \textbf{Labs}\\
    \hline
    \begin{verbatim}
Based on your previous answers and the {lab} measurement codes provided in the 
provided reference file.

suggest normal expected values for the lab test in {region}. 

Please use the description column in the data dictionary to describe the content of 
the column.
Each column/field name can be used for as many lab tests as needed. 
Build a csv table with ALL of the following columns: 'flag', 'column name', 
'codes', 'data type','value column name','description','expected value',
\end{verbatim}\\
\begin{verbatim}
'expected value max', expected value min','range low', 'range high',
'comparison type','references'. 
All columns are mandatory. If a field is irrelevant, fill it with np.nan. Do not 
suggest random population statistics. Only use columns and tables from the provided 
data dictionary.
Use only codes in the reference file.
Please use your previous replies.
The 'flag' column should indicate the data type and should be filled with one of the 
following values: 'demography', 'drug', 'procedure', 'lab' or 'diagnosis'.
'value column name' should contain the field name from the provided data dictionary 
with the lab test result value, if applicable. If it doesn't exist, fill it with 
'np.nan'.
Please make sure the correct field name for 'value column name' is in the output csv.
Before the CSV, include the title: "Title: expected_params_labs". Do not separate it
from the CSV code block.
Use the vertical bar | as the CSV separator.
Surround the CSV table with ```csv and ```.
In the 'column name' field, use the corresponding column name from the data 
dictionary.
For the 'comparison type', specify whether the comparison is an 'exact value', 
'range', or 'percentage'. If it is a range, use numeric values only.
Do not suggest random numbers. Only use columns and tables from the provided data 
dictionary.
In the 'codes' column, insert all the relevant {lab} lab test codes from the 
reference file as a list. 
Each lab test should appear in a separate row. Include the code meaning in the 
'description' column, double check the meaning and provide references for the 
statistical values. 
Keep the 'description' column short, concise and specific with exact descripton of 
what is evaluated.
In the values, include only the number, without percentage or units. Expected values 
should be normal range of values for the evaluated lab test.
Use all of the provided information to complete the CSV.
Provide as many column evaluations from the given data dictionary as possible.
Make sure the 'description' column includes the description of the column from the 
data dictionary, including the meaning of the corresponding {coding} codes. 
Provide references for the code meanings and for the statistical values. Work step 
by step. Give me your best answer.
    \end{verbatim}\\
    \hline
    \textbf{Creation of lab and procedure test suggestion CSVs}\\
    \hline{}
    \begin{verbatim}
Please provide a single and final csv file for labs based on your previous answer. 
Include the required title and wrap it between '```csv' and '```'.
    \end{verbatim}\\
\hline
\begin{verbatim}
Please provide a single and final csv file for procedures based on your previous 
answers. **Make sure there are no duplications with the labs csv**. 
Include the required title and wrap it between '```csv' and '```'.
\end{verbatim}\\
\hline
\textbf{General population demographics}\\
\hline
\begin{verbatim}
Look at the diagnosis-related table and columns in the following data dictionary:

{data_struct}

In the description, what are the available gender classification codes, according to
the description column? Please provide the codes and their meanings.
\end{verbatim}\\
\hline
\begin{verbatim}
Look at the diagnosis-related table and columns in the provided data dictionary.
In the description, what are the available ethnicity classification codes, according
to the description column? Please provide the codes and their meanings.
\end{verbatim}\\
\hline
\begin{verbatim}
Look at the diagnosis-related table and columns in the provided data dictionary.
In the description, what are the available race classification codes, according to 
the description column? Please provide the codes and their meanings.
\end{verbatim}\\
\hline
\begin{verbatim}
According to the data dictionary, your previous answers and the following reference 
file:

{info}

Please use the coding codes and categorizations provided in the reference file and
\end{verbatim}\\
\begin{verbatim}
the data dictionary to suggest expected values for the general population 
demographics in {region}.

For each column/field name, suggest expected percentage values for the **general 
population** in {region}, according to the provided data.
Please use the description column in the data dictionary to describe the content of 
the column.
You can reuse each data field as many times as necessary to represent various 
population characteristics and subpopulations.

Create a CSV table with the following columns: 
'flag','column name','codes','data type','value column name','description',
'expected value','expected value max','expected value min','range low',
'range high','comparison type','references'

Follow these guidelines:
All columns are mandatory. If a field is irrelevant, fill it with np.nan.
Each demographic subpopulation should appear in a separate row (e.g a row for female,
male, white, etc).
The possible values for gender, race and ethnicity are stated in the 'description' 
column in the provided data dictionary for the relevant data field. Please use them 
as well.
Please prefer the codes provided in the data dictionary and only then the codes in
the reference file. You may add all options in the provided data for each 
subpopulation.
The 'flag' column should indicate the data type and should be filled with one of the 
following values: 'demography', 'drug', 'procedure', or 'diagnosis'.
'value column name' should contain the actual name of the column from the provided 
data dictionary that holds the value.
Before the CSV, include the title: "Title: expected_params_genpop_demo". Do not 
separate it from the CSV code block.
Use the vertical bar | as the CSV separator.
Surround the CSV table with ```csv and ```.
In the 'column name' field, use the corresponding column/field name from the data 
dictionary.
For the 'comparison type', specify whether the comparison is an 'exact value', 
'range', or 'percentage'. If it is a range, use numeric values only.
If the provided expected value is a range, fill "range low" and "range high" and set 
'comparison type' to be a 'range'.
\end{verbatim}\\
\begin{verbatim}
Do not suggest random population statistics. Only use columns and tables from the 
provided data dictionary.
In the 'codes' column, for each gender, race and ethnicity subpopulation insert 
relevant classification codes **from the reference file, your previous answers and 
the data dictionary** as a list. 
Add the relevant gender codes, race codes and ethnicity codes from your previous 
answers to the appropriate list under the 'codes' column. The codes that don't have 
their own row should appear in 'other' race, gender or ethnicity row.
Each demographic subpopulation should appear in a separate row (e.g a row for male, a 
row for female, a row for white, a row for asian, etc.). Include the code 
meaning in the 'description' column, double check the meaning and provide references 
for the statistical values. Keep the 'description' column short, concise and specific 
with exact descripton of what is evaluated.
Use all of the provided information to complete the CSV. 
Provide as many column evaluations from the given data dictionary as possible.
For date-related data, prefer columns with datetime values where applicable.
Make sure the 'description' column includes the description of the column from the 
data dictionary, including the meaning of the corresponding {coding} codes. 
If there are any additional demographic {coding} codes that might apply to {region},
please **add** them to the existing list - to check for additional codes you may use 
your intrinsic knowledge.
Provide references for the code meanings and for the statistical values.
double check your reply.
Please make sure there is a separate row for each subpopulation (e.g. for gender - 
separate rows for male and for female, as well as for the different race and 
ethnicity options).
The codes should correspond. Provide your best answer.
\end{verbatim}\\
\hline
\begin{verbatim}
Please provide a single and final csv file for the general population demographics 
based on your previous answers. Include all gender, race, ethnicity and any other 
categorical fields. 
Make sure all relevant gender, race and ethnicity classification codes are included 
in the 'codes' column.
Include the required title and wrap it between '```csv' and '```'.
\end{verbatim}\\
\hline
\begin{verbatim}
Please make sure that the previous csv file contains a separate row for each 
demographic subgroup (e.g a row for female, male, white, etc).
If it does not, please recreate it so it follows the required gudelines.
\end{verbatim}\\
\hline
\textbf{Diagnosed population demographics}\\
\hline
\begin{verbatim}
According to the data dictionary, your previous answers and the provided reference 
file.

Please use the coding codes and categorizations provided in the reference file and 
the data dictionary to suggest expected values for the {diagnosis} diagnosed 
population demographics in {region}.
For each column/field name, suggest expected demographic statistics values for the 
population diagnosed with {diagnosis} in {region}.
Please use the description column in the data dictionary to describe the content of
the column.
Use each column as needed for as many population characteristics as possible. Build a
CSV table with all the following columns:
'flag','column name','codes','data type','value column name','description',
'expected value','expected value max','expected value min','range low',
'range high','comparison type','references'

Follow these guidelines:
All columns are mandatory. If a field is irrelevant, fill it with np.nan.
Each demographic subpopulation should appear in a separate row (e.g a row for female,
male, white, etc).
The possible values for gender, race and ethnicity are stated in the 'description'
column in the provided data dictionary for the relevant data field. Please use them 
as well.
Please prefer the codes provided in the data dictionary and only then the codes in 
the reference file. You may add all options in the provided data for each 
subpopulation.
The 'flag' column should indicate the data type and should be filled with one of the
following values: 'demography', 'drug', 'procedure', 'lab' or 'diagnosis'.
'value column name' should contain the actual name of the column from the provided 
data dictionary that holds the value.
\end{verbatim}\\
\begin{verbatim}
Before the CSV, include the title: "Title: expected_params_diagnosed_genpop". Do not 
separate it from the CSV code block.
Use the vertical bar | as the CSV separator.
Surround the CSV table with ```csv and ```.
In the 'column name' field, use the corresponding column name from the data 
dictionary.
For the 'comparison type', specify whether the comparison is an 'exact value', 
'range', or 'percentage'. If it is a range, use numeric values only.
If the provided expected value is a range, fill "range low" and "range high" and set
'comparison type' to be a 'range'.
Do not suggest random population statistics. Only use columns and tables from the 
provided data dictionary.
In the 'codes' column, for each gender, race and ethnicity subpopulation insert 
relevant classification codes **from the reference file, your previous answers and 
the data dictionary** as a list. 
Add the relevant gender codes, race codes and ethnicity codes from your previous 
answers to the appropriate list under the 'codes' column. The codes that don't have
their own row should appear in 'other' race, gender or ethnicity row.

Each demographic subpopulation should appear separately (e.g a row for male, a row 
for female, a row for white, a row for asian, etc.). Include the code meaning in 
the 'description' column, double check the meaning and provide references for the 
statistical values. Keep the 'description' column short, concise and specific with
exact descripton of what is evaluated - state the exact gender or race or etnicity
evaluated.
In the 'expected value', 'expected value min', 'expected value max' columns, provide
the percentage of **diagnosed population** with the specific demographic 
characteristic.
Use all of the provided information to complete the CSV.
Provide as many column evaluations from the given data dictionary as possible.
For date-related data, prefer columns with datetime values where applicable.
Make sure the 'description' column includes the description of the column from the 
data dictionary, including the meaning of the corresponding {coding} codes. 
If there are any additional demographic {coding} codes that might apply to {region},
please **add** them to the existing list - to check for additional codes you may use 
your intrinsic knowledge.
Provide references for the code meanings and the values. Alwaus double check your 
reply. Always work step by step.
Please create a separate row for each subpopulation (e.g. for gender - separate rows
for male and for female, as well as for the different race and ethnicity options).
\end{verbatim}\\
\hline
\begin{verbatim}
Please provide a single and final csv file for the **diagnosed population 
demographics** based on your previous answers. Include all gender, race, ethnicity 
and any other categorical fields. 
Make sure all relevant gender, race and ethnicity classification codes are included 
in the 'codes' column. 
Include the required title and wrap it between '```csv' and '```'.
\end{verbatim}\\
\hline
\begin{verbatim}
Please make sure that the previous csv file contains a separate row for each 
demographic subgroup (e.g a row for female, male, white, etc).
If it does not, please recreate it so it follows the required gudelines.
\end{verbatim}\\
\hline
\textbf{Demographic code and test validation}\\
\hline
\begin{verbatim}
Please use the following data dictionary to answer the following questions: \n\n +
{data_struct}
\end{verbatim}\\
\hline
\begin{verbatim}
This is a csv containing proposed evaluations: \n\n {messages4[-1]['content']}
\end{verbatim}\\
\hline
\begin{verbatim}
Are there additional relevant codes in one of the rows in the provided table above, 
that correspond with the value under 'description' column in each row?
Please return the same table with the same title, with **additional** codes in each 
row, if applicable. Provide references. 
Do not change anything other than what was requested.
Please work step by step and provide your best answer
\end{verbatim}\\
\hline
\textbf{Disease statistics validation}\\
\hline
\begin{verbatim}
Are there additional relevant codes in one of the rows in the provided table above, 
that correspond with the value under 'description' column in each row? 
Please return the same table with the same title, with additional codes in each row,
if applicable. Provide references. Ignore this request for the 
'convert_to_datastruct', 'expected_params_procedures', 'expected_params_drugs', and
'expected_params_comorbidities' tables.
Please work step by step and provide your best answer.
\end{verbatim}\\
\hline
\begin{verbatim}
Make sure the proposed column names for evaluation correspond with the ones in the 
provided data dictionary and with the requirements in the system message.

If it doesn't correspond to what was reuired, please change the evaluated column 
names according to what was required.
Please work step by step and provide your best answer.
\end{verbatim}\\
\hline
\begin{verbatim}
Make sure the evaluated values correspond with the ones in the provided supplementary
data, data dictionary and the requirements in the first message.
If it doesn't correspond to what was reuired, please change the values according to 
what is evaluated.
Please work step by step, provide references and provide your best answer.
\end{verbatim}\\
\hline
\end{longtable}

\newpage

\begin{table}
\caption{Performance metrics of the Auditor Agent across clinical datasets. The table quantifies the automated refinement of generated tests, distinguishing between corrections (realignment with ground truth) and discards (removal of irreconcilable noise).}
\label{tab:auditor_correct_discard}
\centering
\setlength{\tabcolsep}{4pt}
\begin{tabularx}{\linewidth}{l *{8}{>{\centering\arraybackslash}X}}
\toprule
Dataset & Original & Kept & Corrected & Discarded & Added & Corr. rate (\%) & Disc. rate (\%) & Kept rate (\%) \\
\midrule
CHF & 60 & 20 & 33 & 6 & 0 & 55.00 & 10.00 & 33.30 \\
CKD & 96 & 45 & 43 & 8 & 0 & 44.80 & 8.30 & 46.90 \\
HTN & 78 & 26 & 45 & 7 & 1 & 57.70 & 9.00 & 33.30 \\
T2D & 94 & 45 & 37 & 2 & 0 & 39.40 & 2.10 & 47.90 \\
\bottomrule
\end{tabularx}
\end{table}

\newpage

\setlength\LTcapwidth{15cm}
\begin{longtable}{p{5cm}p{10cm}}
\caption{Comprehensive list of tests generated for the Hypertension (SyntheticMass), CHF (MIMIC-III), CKD (AoU), and T2D (AoU) cohorts.}\\
\label{tab:all_testsֿֿֿֿ}\\
\toprule
Cohort & Test Name \\
\midrule
\endfirsthead

\toprule
Cohort & Test Name \\
\midrule
\endhead

\bottomrule
\endfoot

\bottomrule
\endlastfoot

Hypertension - SyntheticMass & check data types \\
Hypertension - SyntheticMass & check incidence \\
Hypertension - SyntheticMass & check prevalence \\
Hypertension - SyntheticMass & calculate age distribution \\
Hypertension - SyntheticMass & Diabetes mellitus diagnosis code among diagnosed \\
Hypertension - SyntheticMass & Hyperlipidemia diagnosis code among diagnosed \\
Hypertension - SyntheticMass & Coronary artery disease diagnosis code among diagnosed \\
Hypertension - SyntheticMass & Obesity diagnosis code among diagnosed \\
Hypertension - SyntheticMass & Heart failure diagnosis code among diagnosed \\
Hypertension - SyntheticMass & Atrial fibrillation and related conditions diagnosis code among diagnosed \\
Hypertension - SyntheticMass & Stroke and related conditions diagnosis code among diagnosed \\
Hypertension - SyntheticMass & Peripheral vascular disease diagnosis code among diagnosed \\
Hypertension - SyntheticMass & Chronic obstructive pulmonary disease diagnosis code among diagnosed \\
Hypertension - SyntheticMass & Sleep apnea and related conditions diagnosis code among diagnosed \\
Hypertension - SyntheticMass & Depression diagnosis code among diagnosed \\
Hypertension - SyntheticMass & Osteoarthritis diagnosis code among diagnosed \\
Hypertension - SyntheticMass & Anemia diagnosis code among diagnosed \\
Hypertension - SyntheticMass & Male gender \\
Hypertension - SyntheticMass & Female gender \\
Hypertension - SyntheticMass & White race \\
Hypertension - SyntheticMass & Black or African American race \\
Hypertension - SyntheticMass & Asian race \\
Hypertension - SyntheticMass & American Indian or Alaska Native race \\
Hypertension - SyntheticMass & Native Hawaiian or Other Pacific Islander race \\
Hypertension - SyntheticMass & Other race \\
Hypertension - SyntheticMass & Hispanic or Latino ethnicity \\
Hypertension - SyntheticMass & Not Hispanic or Latino ethnicity \\
Hypertension - SyntheticMass & Lisinopril \\
Hypertension - SyntheticMass & Amlodipine \\
Hypertension - SyntheticMass & Losartan \\
Hypertension - SyntheticMass & Hydrochlorothiazide \\
Hypertension - SyntheticMass & Metoprolol \\
Hypertension - SyntheticMass & Atenolol \\
Hypertension - SyntheticMass & Valsartan \\
Hypertension - SyntheticMass & Enalapril \\
Hypertension - SyntheticMass & Carvedilol \\
Hypertension - SyntheticMass & Furosemide \\
Hypertension - SyntheticMass & Clonidine \\
Hypertension - SyntheticMass & Male gender diagnosed with hypertension \\
Hypertension - SyntheticMass & Female gender diagnosed with hypertension \\
Hypertension - SyntheticMass & White race diagnosed with hypertension \\
Hypertension - SyntheticMass & Black or African American race diagnosed with hypertension \\
Hypertension - SyntheticMass & Asian race diagnosed with hypertension \\
Hypertension - SyntheticMass & American Indian or Alaska Native race diagnosed with hypertension \\
Hypertension - SyntheticMass & Native Hawaiian or Other Pacific Islander race diagnosed with hypertension \\
Hypertension - SyntheticMass & Hispanic or Latino ethnicity diagnosed with hypertension \\
Hypertension - SyntheticMass & Not Hispanic or Latino ethnicity diagnosed with hypertension \\
Hypertension - SyntheticMass & Blood pressure measurement \\
Hypertension - SyntheticMass & Lipid panel measurement \\
Hypertension - SyntheticMass & Serum glucose level \\
Hypertension - SyntheticMass & Serum cholesterol level \\
Hypertension - SyntheticMass & Serum HDL cholesterol level \\
Hypertension - SyntheticMass & Serum LDL cholesterol level \\
Hypertension - SyntheticMass & Serum triglycerides level \\
Hypertension - SyntheticMass & Serum potassium level \\
Hypertension - SyntheticMass & Serum creatinine level \\
Hypertension - SyntheticMass & Serum sodium level \\
Hypertension - SyntheticMass & Serum calcium level \\
Hypertension - SyntheticMass & Serum magnesium level \\
Hypertension - SyntheticMass & Serum chloride level \\
Hypertension - SyntheticMass & Serum bicarbonate level \\
Hypertension - SyntheticMass & Serum urea nitrogen level \\
Hypertension - SyntheticMass & Serum uric acid level \\
Hypertension - SyntheticMass & Serum total protein level \\
Hypertension - SyntheticMass & Serum albumin level \\
Hypertension - SyntheticMass & Serum globulin level \\
Hypertension - SyntheticMass & Serum bilirubin level \\
Hypertension - SyntheticMass & Serum alkaline phosphatase level \\
Hypertension - SyntheticMass & Serum aspartate aminotransferase level \\
Hypertension - SyntheticMass & Serum alanine aminotransferase level \\
Hypertension - SyntheticMass & Serum gamma glutamyl transferase level \\
Hypertension - SyntheticMass & Serum lactate dehydrogenase level \\
Hypertension - SyntheticMass & Serum C reactive protein level \\
CHF - MIMIC-III & check data types \\
CHF - MIMIC-III & check incidence \\
CHF - MIMIC-III & check prevalence \\
CHF - MIMIC-III & calculate age distribution \\
CHF - MIMIC-III & Hypertension diagnosis codes among diagnosed \\
CHF - MIMIC-III & Diabetes Mellitus diagnosis codes among diagnosed \\
CHF - MIMIC-III & Chronic Obstructive Pulmonary Disease diagnosis codes among diagnosed \\
CHF - MIMIC-III & Atrial Fibrillation diagnosis codes among diagnosed \\
CHF - MIMIC-III & Coronary Artery Disease diagnosis codes among diagnosed \\
CHF - MIMIC-III & Anemia diagnosis codes among diagnosed \\
CHF - MIMIC-III & Renal Failure diagnosis codes among diagnosed \\
CHF - MIMIC-III & Obesity diagnosis codes among diagnosed \\
CHF - MIMIC-III & Hyperlipidemia diagnosis codes among diagnosed \\
CHF - MIMIC-III & Depression diagnosis codes among diagnosed \\
CHF - MIMIC-III & Peripheral Vascular Disease diagnosis codes among diagnosed \\
CHF - MIMIC-III & Sleep Apnea diagnosis codes among diagnosed \\
CHF - MIMIC-III & Hypothyroidism diagnosis codes among diagnosed \\
CHF - MIMIC-III & Stroke diagnosis codes among diagnosed \\
CHF - MIMIC-III & Liver Disease diagnosis codes among diagnosed \\
CHF - MIMIC-III & Male gender \\
CHF - MIMIC-III & Female gender \\
CHF - MIMIC-III & White race \\
CHF - MIMIC-III & Black or African American race \\
CHF - MIMIC-III & Asian race \\
CHF - MIMIC-III & Hispanic Latino ethnicity \\
CHF - MIMIC-III & Other ethnicity \\
CHF - MIMIC-III & Lisinopril \\
CHF - MIMIC-III & Carvedilol \\
CHF - MIMIC-III & Furosemide \\
CHF - MIMIC-III & Spironolactone \\
CHF - MIMIC-III & Digoxin \\
CHF - MIMIC-III & Valsartan \\
CHF - MIMIC-III & Sacubitril Valsartan \\
CHF - MIMIC-III & Bisoprolol \\
CHF - MIMIC-III & Ivabradine \\
CHF - MIMIC-III & Eplerenone \\
CHF - MIMIC-III & Torsemide \\
CHF - MIMIC-III & Male gender diagnosed with CHF \\
CHF - MIMIC-III & Black or African American race diagnosed with CHF \\
CHF - MIMIC-III & Asian race diagnosed with CHF \\
CHF - MIMIC-III & Hispanic Latino ethnicity diagnosed with CHF \\
CHF - MIMIC-III & Other ethnicity diagnosed with CHF \\
CHF - MIMIC-III & Radiologic examination chest single view \\
CHF - MIMIC-III & Transferase aspartate amino AST SGOT \\
CHF - MIMIC-III & Alanine aminotransferase ALT measurement \\
CHF - MIMIC-III & B type Natriuretic Peptide BNP measurement \\
CHF - MIMIC-III & Troponin I measurement \\
CHF - MIMIC-III & Troponin T measurement \\
CHF - MIMIC-III & Serum Creatinine measurement \\
CHF - MIMIC-III & Blood Urea Nitrogen BUN measurement \\
CHF - MIMIC-III & Serum Electrolytes measurement \\
CHF - MIMIC-III & Serum Magnesium measurement \\
CHF - MIMIC-III & Serum Calcium measurement \\
CHF - MIMIC-III & Serum Phosphate measurement \\
CHF - MIMIC-III & Serum Glucose measurement \\
CKD - All of Us & check data types \\
CKD - All of Us & check incidence \\
CKD - All of Us & check prevalence \\
CKD - All of Us & calculate age distribution \\
CKD - All of Us & Hypertension diagnosis codes among diagnosed \\
CKD - All of Us & Heart failure diagnosis codes among diagnosed \\
CKD - All of Us & Diabetes Mellitus without complication diagnosis codes among diagnosed \\
CKD - All of Us & Diabetes Mellitus with complication diagnosis codes among diagnosed \\
CKD - All of Us & Hyperlipidemia diagnosis codes among diagnosed \\
CKD - All of Us & Coronary Artery Disease diagnosis codes among diagnosed \\
CKD - All of Us & Peripheral Vascular Disease diagnosis codes among diagnosed \\
CKD - All of Us & Depression diagnosis codes among diagnosed \\
CKD - All of Us & Osteoporosis diagnosis codes among diagnosed \\
CKD - All of Us & Osteopenia diagnosis codes among diagnosed \\
CKD - All of Us & Gout diagnosis codes among diagnosed \\
CKD - All of Us & Sleep Apnea diagnosis codes among diagnosed \\
CKD - All of Us & Hyperparathyroidism diagnosis codes among diagnosed \\
CKD - All of Us & Liver Disease diagnosis codes among diagnosed \\
CKD - All of Us & Female gender \\
CKD - All of Us & Male gender \\
CKD - All of Us & White race \\
CKD - All of Us & Black or African American race \\
CKD - All of Us & Asian race \\
CKD - All of Us & Native Hawaiian or Other Pacific Islander race \\
CKD - All of Us & American Indian or Alaska Native race \\
CKD - All of Us & Not Hispanic or Latino ethnicity \\
CKD - All of Us & Hispanic or Latino ethnicity \\
CKD - All of Us & Lisinopril \\
CKD - All of Us & Amlodipine \\
CKD - All of Us & Metoprolol \\
CKD - All of Us & Furosemide \\
CKD - All of Us & Erythropoietin \\
CKD - All of Us & Calcitriol \\
CKD - All of Us & Cinacalcet \\
CKD - All of Us & Female gender diagnosed with CKD \\
CKD - All of Us & White race diagnosed with CKD \\
CKD - All of Us & Black or African American race diagnosed with CKD \\
CKD - All of Us & Asian race diagnosed with CKD \\
CKD - All of Us & Native Hawaiian or Other Pacific Islander race diagnosed with CKD \\
CKD - All of Us & American Indian or Alaska Native race diagnosed with CKD \\
CKD - All of Us & Not Hispanic or Latino ethnicity diagnosed with CKD \\
CKD - All of Us & Hispanic or Latino ethnicity diagnosed with CKD \\
CKD - All of Us & Serum Creatinine Measurement \\
CKD - All of Us & Blood Urea Nitrogen Measurement \\
CKD - All of Us & Serum Potassium Measurement \\
CKD - All of Us & Serum Sodium Measurement \\
CKD - All of Us & Serum Calcium Measurement \\
CKD - All of Us & Serum Phosphorus Measurement \\
CKD - All of Us & Serum Albumin Measurement \\
CKD - All of Us & Hemoglobin Measurement \\
CKD - All of Us & Hematocrit Measurement \\
CKD - All of Us & Glomerular Filtration Rate Measurement \\
CKD - All of Us & Urine Protein Measurement \\
CKD - All of Us & Urine Albumin Measurement \\
CKD - All of Us & Urine Creatinine Measurement \\
CKD - All of Us & Parathyroid Hormone Measurement \\
CKD - All of Us & Serum Bicarbonate Measurement \\
CKD - All of Us & Serum Chloride Measurement \\
CKD - All of Us & Serum Magnesium Measurement \\
CKD - All of Us & Serum Uric Acid Measurement \\
CKD - All of Us & Serum Total Protein Measurement \\
CKD - All of Us & Serum Alkaline Phosphatase Measurement \\
CKD - All of Us & Serum Aspartate Aminotransferase Measurement \\
CKD - All of Us & Serum Alanine Aminotransferase Measurement \\
CKD - All of Us & Serum Bilirubin Measurement \\
CKD - All of Us & Serum Ferritin Measurement \\
CKD - All of Us & Serum Transferrin Measurement \\
T2D - All of Us & check data types \\
T2D - All of Us & check incidence \\
T2D - All of Us & check prevalence \\
T2D - All of Us & calculate age distribution \\
T2D - All of Us & Hypertension diagnosis among diagnosed \\
T2D - All of Us & Hyperlipidemia diagnosis among diagnosed \\
T2D - All of Us & Coronary artery disease diagnosis among diagnosed \\
T2D - All of Us & Chronic kidney disease diagnosis among diagnosed \\
T2D - All of Us & Peripheral vascular disease diagnosis among diagnosed \\
T2D - All of Us & Retinopathy diagnosis among diagnosed \\
T2D - All of Us & Stroke diagnosis among diagnosed \\
T2D - All of Us & Depression diagnosis among diagnosed \\
T2D - All of Us & Osteoarthritis diagnosis among diagnosed \\
T2D - All of Us & Sleep apnea diagnosis among diagnosed \\
T2D - All of Us & Fatty liver disease diagnosis among diagnosed \\
T2D - All of Us & Anemia diagnosis among diagnosed \\
T2D - All of Us & Female gender \\
T2D - All of Us & Male gender \\
T2D - All of Us & White race \\
T2D - All of Us & Black or African American race \\
T2D - All of Us & Asian race \\
T2D - All of Us & Native Hawaiian or Other Pacific Islander race \\
T2D - All of Us & American Indian or Alaska Native race \\
T2D - All of Us & Other race \\
T2D - All of Us & Not Hispanic or Latino ethnicity \\
T2D - All of Us & Hispanic or Latino ethnicity \\
T2D - All of Us & Metformin \\
T2D - All of Us & Sitagliptin \\
T2D - All of Us & Saxagliptin \\
T2D - All of Us & Male gender diagnosed with T2D \\
T2D - All of Us & White race diagnosed with T2D \\
T2D - All of Us & Black or African American race diagnosed with T2D \\
T2D - All of Us & Asian race diagnosed with T2D \\
T2D - All of Us & Native Hawaiian or Other Pacific Islander race diagnosed with T2D \\
T2D - All of Us & Not Hispanic or Latino ethnicity diagnosed with T2D \\
T2D - All of Us & Hispanic or Latino ethnicity diagnosed with T2D \\
T2D - All of Us & Hemoglobin A1c measurement \\
T2D - All of Us & Serum glucose measurement \\
T2D - All of Us & Urine microalbumin measurement \\
T2D - All of Us & Serum creatinine measurement \\
T2D - All of Us & Lipid panel \\
T2D - All of Us & Blood urea nitrogen measurement \\
T2D - All of Us & Serum potassium measurement \\
T2D - All of Us & Serum sodium measurement \\
T2D - All of Us & Serum chloride measurement \\
T2D - All of Us & Serum bicarbonate measurement \\
T2D - All of Us & Serum calcium measurement \\
T2D - All of Us & Serum magnesium measurement \\
T2D - All of Us & Serum albumin measurement \\
T2D - All of Us & Serum bilirubin measurement \\
T2D - All of Us & Serum alkaline phosphatase measurement \\
T2D - All of Us & Serum aspartate aminotransferase measurement \\
T2D - All of Us & Serum alanine aminotransferase measurement \\
T2D - All of Us & Serum lactate dehydrogenase measurement \\
T2D - All of Us & Serum uric acid measurement \\
T2D - All of Us & Serum triglycerides measurement \\
T2D - All of Us & Serum high density lipoprotein measurement \\
T2D - All of Us & Serum low density lipoprotein measurement \\
T2D - All of Us & Serum phosphate measurement \\
T2D - All of Us & Serum cholesterol measurement \\
\bottomrule
\end{longtable}

\newpage

\begin{longtable}{p{15cm}}
\caption{Test suggestion validation prompts}\\
    \label{tab:test_val_prompts}\\
    \hline
       \textbf{Validation of test suggestion tables} \\
    \hline
       \begin{verbatim}
For the following test, please evaluate the reference value by searching for the
correct reference value. 
It appears as the expected value or range. Please make sure that the reference value
is correct. 
If there is no available data for {region}, please provide the reference value for
the whole country, continent or global statistics.
If it is incorrect, please provide the correct value. If you are unsure, please 
indicate that as well. 
For disease-related tests, the reference value represents the prevalence of that 
disease among patients with {diagnosis} in {region}.
Demographics - unless diagnosed is 'True', the percentage is relative to the general
population in {region}. 
For those diagnosed with {diagnosis} in {region}, diagnosed flag will be 'True', and
the percentage is the percentage of {diagnosis} patients among the evaluated 
\end{verbatim}\\
\begin{verbatim}
statistic (e.g, percentage of diagnosed women among all women or percentage of
patients prescribed a drug among all patients with {diagnosis}).

The test suggestion is as follows:

Column names: {list(row.index)}
Values: {row.values}

diagnosed: {diagnosed}

Please provide your feedback in the following format:
1. Test description: [your answer]
2. Correct reference value: [your answer]
3. Provided reference value: [your answer]
4. Feedback: [your answer - Correct or Incorrect or Unsure]
5. Fuction recommendation: [your answer - Fix (if incorrect), No fix (if correct), Remove (if unsure)]
\end{verbatim}\\
\hline
\begin{verbatim}
If according to your feedback the test suggestion needs fixing due to reference 
value correction, please provide it betweeen triple backticks (```csv). use a 
vertical line | as separator.
do not change anything but the reference value and the comparison type, if needed.
Strictly stick to the original field names and order. To the field names add 'fixed'
and 'remove'.
Set 'fixed' to be 'True' to the test suggestion if the function needs to be fixed. 
Set it to be 'False' otherwise.   
Make sure the fixed value corrersponds with the correct value you found.
If the expected value is a range and not a single value, please fill the range_high 
and range_low columns with the correct values, and remove the incorrect value. In 
that case - change the comparison_type to 'range'.
Please make sure the numbers are in the correct location (minimum and maximum or 
high and low).
The comparison_type can also be set to 'less_than' or 'greater_than', in there is a
minimum or a maximum only value, respectively.
If you recommended removing the test, please set the 'remove' column in the test 
suggestion to be 'True'. No need to fix it in that case. 
If the column needs to be fixed, fix it, set 'fixed' to be 'True' and set 'remove'
column to be 'False'.
\end{verbatim}\\
\begin{verbatim}
If a test repeats - in terms of 'description' and 'expected_value' or range columns,
please set the 'remove' column to be 'True' for the repeated test suggestion.
Please add the following references (including urls) to the 'references' column in 
the corrected csv. Add all the values in the 'references' column in parentheses "".
                         
     {replies[-1][ref_loc:]}

Return the test anyway, even if it is correct, with the additional columns (both set
to 'False', if fix or removal isn't suggested).
\end{verbatim}\\
\hline
\textbf{Validation of statistical values}\\
\hline
\begin{verbatim}
For the following statistics, please evaluate the reference value by searching for 
the correct reference value. 
It appears as the expected value or range. Please make sure that the reference value
is correct. 
If there is no available data for {region}, please provide the reference value for 
the whole country, continent or global statistics.
If it is incorrect, please provide the correct value. If you are unsure, please 
indicate that as well. 

The statistics are for {diagnosis} in {region}:

{stats}

Please provide your feedback in the following format for each statistic:
1. Statistic: [your answer]
2. Correct reference value: [your answer]
3. Provided reference value: [your answer]
4. Feedback: [your answer - Correct or Incorrect or Unsure]
5. Recommendation: [your answer - Fix (if incorrect), No fix (if correct), Remove 
(if unsure)]
\end{verbatim}\\
\hline
\begin{verbatim}
Please provide the modified statistics in a similar format as the original, betweeen
triple backticks (```csv). 
use a vertical line | as separator.
If according to your feedback the statitic suggestion needs fixing due to reference
value correction, please correct it.
do not change anything but the reference value.
Strictly stick to the original field names and order.
Make sure the fixed value corrersponds with the correct value you found.
If you recommended removing the statistic, replace the relevant statistic with 
'np.nan'. 
Please add the following references (including urls) to the 'reference' column in 
the corrected csv. Add all the values in the 'reference' column in parentheses "".
                
{replies[-1][ref_loc:]}
\end{verbatim}\\
\hline
\begin{verbatim}
For the following statistics, please evaluate the reference value by searching for 
the correct reference value. 
It appears as the expected value or range. Please make sure that the reference value
is correct. 
If there is no available data for {region}, please provide the reference value for 
the whole country, continent or global statistics.
If it is incorrect, please provide the correct value. If you are unsure, please 
indicate that as well. 

The statistics are for {diagnosis} in {region}:

{age}

Please provide your feedback in the following format for each statistic:
1. Statistic: [your answer]
2. Correct reference value: [your answer]
3. Provided reference value: [your answer]
4. Feedback: [your answer - Correct or Incorrect or Unsure]
5. Recommendation: [your answer - Fix (if incorrect), No fix (if correct), Remove
(if unsure)]
\end{verbatim}\\
\hline
\begin{verbatim}
Please provide the modified statistics in a similar format as the original, betweeen
triple backticks (```csv). 
use a vertical line | as separator.
If according to your feedback the statitic suggestion needs fixing due to reference 
value correction, please correct it.
do not change anything but the reference value.
Strictly stick to the original field names and order.
Make sure the fixed value corrersponds with the correct value you found.
If you recommended removing the statistic, replace the relevant statistic with 
'np.nan'. 
Please add the following references (including urls) to the 'reference' column in 
the corrected csv. Add all the values in the 'reference' column in parentheses "".

{replies[-1][ref_loc:]}
\end{verbatim}\\
\hline

\end{longtable}

\begin{table}
\footnotesize

    \begin{tabular}{lcccccccc}
    \hline
        & 
         & \textbf{Valid References}
          & \textbf{Total Tests}
        & \textbf{Accurate} 
        & \textbf{Inaccurate}
        & \textbf{Incorrect}
        & \textbf{Qualitative} 
        & \textbf{Irrelevant} \\

         \hline
      & \textbf{\ttd{}}  & 92.9\% &  57 &  62.5\% & 17.9\% & 19.6\% & 0\% & 0\%  \\
                    \hline
      & \textbf{\ckd{}}  & 49.2\% & 60 & 65.0\%  & 16.7\%  & 8.3\%  & 3.4\%  & 5.0\% \\
                    \hline
       & \textbf{\htn{}}  & 98.3\% & 61  & 68.9\%  & 6.6\% &  14.8\%  & 4.9\%  &  3.3\%  \\
         \hline
      & \textbf{\chf{}} & 39.5\% &  44  & 47.7\%  &  25.0\%  &  15.9\% &  2.3\%  & 6.8\%  \\
                        \hline
    \end{tabular}
    \caption{Test quality validation statistics when tests that evaluate prevalence of drug usage are disregarded.}
    \label{tab:test_quality_no_drugs}
\end{table}

\begin{table}
    \centering
    \begin{tabular}{llrrr}
        \toprule
        \cmidrule(lr){1-3}
        & & Population Characteristics \\
        \midrule
        n & & 410,361 \\
        \midrule
        \multirow[t]{4}{*}{\textbf{Gender, n (\%)}} 
         & Female & 245,601 (59.8) \\
         & Male  & 152,142 (37.1) \\
         & No matching concept & 97 (0.0) \\
         & Other / Skipped  & 12,521 (3.1) \\
        \midrule
        \multirow[t]{9}{*}{\textbf{Race, n (\%)}} 
         & Another single population & 2,987 (0.7) \\
         & Asian & 14,162 (3.5) \\
         & Black or African American  & 78,321 (19.1) \\
         & Prefer not to answer  & 2,537 (0.6) \\
         & More than one population  & 7,720 (1.9) \\
         & None indicated  & 64,614 (15.7) \\
         & None of these  & 4,351 (1.1) \\
         & Skip  & 9,054 (2.2) \\
         & White & 226,615 (55.2) \\
        \midrule
        \multirow[t]{6}{*}{\textbf{Ethnicity, n (\%)}} 
         & Hispanic or Latino & 73,945 (18.0) \\
         & No matching concept & 97 (0.0) \\
         & Not Hispanic or Latino & 320,377 (78.1) \\
         & Prefer not to answer & 2,537 (0.6) \\
         & Skip & 9,054 (2.2) \\
         & None of these & 4,351 (1.1) \\
        \midrule
        \multirow[t]{4}{*}{\textbf{Sex at Birth, n (\%)}} 
         & Female & 248,130 (60.5) \\
         & Male & 153,582 (37.4) \\
         & No matching concept & 4,283 (1.0) \\
         & Other / Skipped & 4,366 (1.1) \\
        \midrule
        \textbf{Age, mean (SD)} & & 56.0 (16.8) \\
        \bottomrule
    \end{tabular}
    \caption{\aou{} descriptive statistics.}
    \label{tab:TableOne_AoU}
\end{table}

\begin{table}
\small
    \centering
    \begin{tabular}{lcccc}
\textbf{Test} & \textbf{\ttd{}} & \textbf{\chf{}} & \textbf{\ckd{}} & \textbf{\htn{}} \\
\toprule
\midrule
\textbf{Total Functions} & 60 & 55 & 67 & 73 \\
Total Functions Passed & 6 & 4 & 2 & 2 \\
Total Functions Failed & 54 & 51 & 65 & 71 \\
Total Functions Correct & 35 & 23 & 40 & 43 \\
Total Functions Non-correct & 24 & 31 & 26 & 29 \\
Correct and Passed & 1 & 3 & 0 & 0 \\
Correct and Failed & 34 & 20 & 40 & 43 \\
Non-correct and Passed & 5 & 1 & 2 & 2 \\
Non-correct and Failed & 19 & 30 & 24 & 27 \\
\hline
\textbf{Metadata Total} & 1 & 1 & 1 & 1 \\
Metadata pass & 0 & 0 & 0 & 0 \\
Metadata fail & 1 & 1 & 1 & 1 \\
\hline
\textbf{Subpopulation Total }& 19 & 20 & 22 & 22 \\
Subpopulation Correct & 6 & 3 & 9 & 9 \\
Subpopulation Non-correct & 13 & 17 & 13 & 13 \\
Subpopulation Passed & 2 & 0 & 0 & 1 \\
Subpopulation Failed & 17 & 20 & 22 & 21 \\
Subpopulation Correct and Passed & 0 & 0 & 0 & 0 \\
Subpopulation Correct and Failed & 6 & 3 & 9 & 9 \\
Subpopulation Non-correct and Passed & 2 & 0 & 0 & 1 \\
Subpopulation Non-correct and Failed & 11 & 17 & 13 & 12 \\
\hline
\textbf{Distribution Total} & 40 & 34 & 44 & 50 \\
Distribution Passed & 4 & 4 & 2 & 1 \\
Distribution Failed & 36 & 30 & 42 & 49 \\
Distribution Correct & 29 & 20 & 31 & 32 \\
Distribution Non-correct & 11 & 14 & 13 & 18 \\
Distribution Correct and Passed & 1 & 3 & 0 & 0 \\
Distribution Correct and Failed & 28 & 17 & 31 & 32 \\
Distribution Non-correct and Passed & 3 & 1 & 2 & 1 \\
Distribution Non-correct and Failed & 8 & 13 & 11 & 17 \\
\hline
\textbf{Total Non-correct Functions} & 24 & 31 & 26 & 29 \\
Total inaccurate & 10 & 12 & 12 & 5 \\
Inaccurate and Passed & 3 & 1 & 2 & 1 \\
Inaccurate and Failed & 7 & 11 & 10 & 4 \\
Total incorrect & 12 & 11 & 8 & 19 \\
Incorrect and Passed & 1 & 0 & 0 & 1 \\
Incorrect and Failed & 11 & 11 & 8 & 18 \\
Total irrelevant & 2 & 8 & 6 & 5 \\
Irrelevant and Passed & 1 & 0 & 0 & 0 \\
Irrelevant and Failed & 1 & 8 & 6 & 5 \\
\bottomrule
\end{tabular}
    \caption{Full results describing \mdpt{} performance on the four datasets, in terms of pass/fail and reference correctness.}
    \label{tab:full_results}
\end{table}

\begin{longtable}{lp{4cm}p{7.5cm}}
\caption{Failed Tests with Zero Prevalence} \label{tab:zero_results} \\
\toprule
Cohort & Subcategory & Test Name \\
\midrule
\endfirsthead
\caption[]{Failed Tests with Zero Prevalence} \\
\toprule
Cohort & Subcategory & Test Name \\
\midrule
\endhead
\midrule
\multicolumn{3}{r}{Continued on next page} \\
\midrule
\endfoot
\bottomrule
\endlastfoot
CHF & Demographics (Race/Ethnicity/Gender) & Ethnicity of the patient Asian \\
CHF & Demographics (Race/Ethnicity/Gender) & Ethnicity of the patient Black African American \\
CHF & Demographics (Race/Ethnicity/Gender) & Ethnicity of the patient Hispanic Latino \\
CHF & Demographics (Race/Ethnicity/Gender) & Ethnicity of the patient Other \\
CHF & Demographics (Race/Ethnicity/Gender) & Gender of the patient Male \\
CHF & Medications / Drugs & Bisoprolol \\
CHF & Medications / Drugs & Carvedilol \\
CHF & Medications / Drugs & Digoxin \\
CHF & Medications / Drugs & Eplerenone \\
CHF & Medications / Drugs & Furosemide \\
CHF & Medications / Drugs & Ivabradine \\
CHF & Medications / Drugs & Lisinopril \\
CHF & Medications / Drugs & Sacubitril\_Valsartan \\
CHF & Medications / Drugs & Spironolactone \\
CHF & Medications / Drugs & Torsemide \\
CHF & Medications / Drugs & Valsartan \\
CHF & Other / General Checks & check\_incidence \\
CHF & Other / General Checks & check\_prevalence \\
CKD & Demographics (Race/Ethnicity/Gender) & Race American Indian or Alaska Native \\
CKD & Demographics (Race/Ethnicity/Gender) & Race Asian \\
CKD & Demographics (Race/Ethnicity/Gender) & Race Native Hawaiian or Other Pacific Islander \\
CKD & Diagnoses / Conditions & Coronary Artery Disease diagnosis codes among diagnosed patients \\
CKD & Diagnoses / Conditions & Depression diagnosis codes among diagnosed patients \\
CKD & Diagnoses / Conditions & Diabetes Mellitus with complication diagnosis codes among diagnosed patients \\
CKD & Diagnoses / Conditions & Diabetes Mellitus without complication diagnosis codes among diagnosed patients \\
CKD & Diagnoses / Conditions & Gout diagnosis codes among diagnosed patients \\
CKD & Diagnoses / Conditions & Hyperparathyroidism diagnosis codes among diagnosed patients \\
CKD & Diagnoses / Conditions & Hypertension diagnosis codes among diagnosed patients \\
CKD & Diagnoses / Conditions & Liver Disease diagnosis codes among diagnosed patients \\
CKD & Diagnoses / Conditions & Osteopenia diagnosis codes among diagnosed patients \\
CKD & Diagnoses / Conditions & Osteoporosis diagnosis codes among diagnosed patients \\
CKD & Diagnoses / Conditions & Sleep Apnea diagnosis codes among diagnosed patients \\
CKD & Medications / Drugs & Amlodipine \\
CKD & Medications / Drugs & Calcitriol \\
CKD & Medications / Drugs & Cinacalcet \\
CKD & Medications / Drugs & Erythropoietin \\
CKD & Medications / Drugs & Furosemide \\
CKD & Medications / Drugs & Lisinopril \\
CKD & Medications / Drugs & Metoprolol \\
HTN & Demographics (Race/Ethnicity/Gender) & Hispanic or Latino ethnicity \\
HTN & Demographics (Race/Ethnicity/Gender) & Native Hawaiian or Other Pacific Islander race \\
HTN & Demographics (Race/Ethnicity/Gender) & Not Hispanic or Latino ethnicity \\
HTN & Demographics (Race/Ethnicity/Gender) & Other race \\
HTN & Diagnoses / Conditions & Anemia diagnosis code among diagnosed patients \\
HTN & Diagnoses / Conditions & Atrial fibrillation and related conditions diagnosis code among diagnosed patients \\
HTN & Diagnoses / Conditions & Chronic obstructive pulmonary disease diagnosis code among diagnosed patients \\
HTN & Diagnoses / Conditions & Depression diagnosis code among diagnosed patients \\
HTN & Diagnoses / Conditions & Heart failure diagnosis code among diagnosed patients \\
HTN & Diagnoses / Conditions & Hyperlipidemia diagnosis code among diagnosed patients \\
HTN & Diagnoses / Conditions & Peripheral vascular disease diagnosis code among diagnosed patients \\
HTN & Diagnoses / Conditions & Sleep apnea and related conditions diagnosis code among diagnosed patients \\
HTN & Diagnoses / Conditions & Stroke and related conditions diagnosis code among diagnosed patients \\
HTN & Medications / Drugs & Amlodipine \\
HTN & Medications / Drugs & Atenolol \\
HTN & Medications / Drugs & Carvedilol \\
HTN & Medications / Drugs & Enalapril \\
HTN & Medications / Drugs & Furosemide \\
HTN & Medications / Drugs & Hydrochlorothiazide \\
HTN & Medications / Drugs & Lisinopril \\
HTN & Medications / Drugs & Losartan \\
HTN & Medications / Drugs & Metoprolol \\
HTN & Medications / Drugs & Valsartan \\
T2D & Demographics (Race/Ethnicity/Gender) & American Indian or Alaska Native \\
T2D & Demographics (Race/Ethnicity/Gender) & Asian \\
T2D & Demographics (Race/Ethnicity/Gender) & Native Hawaiian or Other Pacific Islander \\
T2D & Diagnoses / Conditions & Anemia diagnosis code among diagnosed patients \\
T2D & Diagnoses / Conditions & Coronary artery disease diagnosis code among diagnosed patients \\
T2D & Diagnoses / Conditions & Depression diagnosis code among diagnosed patients \\
T2D & Diagnoses / Conditions & Hypertension diagnosis code among diagnosed patients \\
T2D & Diagnoses / Conditions & Osteoarthritis diagnosis code among diagnosed patients \\
T2D & Diagnoses / Conditions & Sleep apnea diagnosis code among diagnosed patients \\
T2D & Diagnoses / Conditions & Stroke diagnosis code among diagnosed patients \\
T2D & Medications / Drugs & Metformin \\
T2D & Medications / Drugs & Sitagliptin \\
T2D & Other / General Checks & Other \\
\end{longtable}

\setlength{\tabcolsep}{5pt}
\begin{longtable}{p{3cm}p{1.2cm}p{1cm}p{2cm}p{2cm}p{1.5cm}p{2cm}p{1.5cm}}
\caption{Sensitivity Analysis Results}\label{tab:sensitivity_results}\\
\toprule
 Cohort & Number of Tests & SMD  & Proportional Tolerance & Fail Number (All) & Fail Rate (All) & Fail Number (Correct) & Fail Rate (Correct) \\
\\midrule
\endfirsthead
\caption[]{Sensitivity Analysis Results} \\
\toprule
 Cohort & Number of Tests & SMD  & Proportional Tolerance & Fail Number (All) & Fail Rate (All) & Fail Number (Correct) & Fail Rate (Correct) \\
\midrule
\endhead
\midrule
\multicolumn{8}{r}{Continued on next page} \\
\midrule
\endfoot
\bottomrule
\endlastfoot
\multirow[t]{15}{3cm}{Congestive Heart Failure (MIMIC-III)} & \multirow[t]{15}{*}{55} & \multirow[t]{5}{*}{0.2} & 0.10 & 53 & 0.964 & 22 & 0.957 \\
 &  &  & 0.15 & 51 & 0.927 & 20 & 0.870 \\
 &  &  & 0.20 & 50 & 0.909 & 19 & 0.826 \\
 &  &  & 0.25 & 50 & 0.909 & 19 & 0.826 \\
 &  &  & 0.30 & 50 & 0.909 & 19 & 0.826 \\
 \hline
 &  & \multirow[t]{5}{*}{0.5} & 0.10 & 53 & 0.964 & 22 & 0.957 \\
 &  &  & 0.15 & 51 & 0.927 & 20 & 0.870 \\
 &  &  & 0.20 & 50 & 0.909 & 19 & 0.826 \\
 &  &  & 0.25 & 50 & 0.909 & 19 & 0.826 \\
 &  &  & 0.30 & 50 & 0.909 & 19 & 0.826 \\
 \hline
 &  & \multirow[t]{5}{*}{0.8} & 0.10 & 53 & 0.964 & 22 & 0.957 \\
 &  &  & 0.15 & 51 & 0.927 & 20 & 0.870 \\
 &  &  & 0.20 & 50 & 0.909 & 19 & 0.826 \\
 &  &  & 0.25 & 50 & 0.909 & 19 & 0.826 \\
 &  &  & 0.30 & 50 & 0.909 & 19 & 0.826 \\
\cline{1-8} \cline{2-8}
\multirow[t]{15}{3cm}{Hypertension (SyntheticMass)} & \multirow[t]{15}{*}{73} & \multirow[t]{5}{*}{0.2} & 0.10 & 72 & 0.986 & 43 & 1.000 \\
 &  &  & 0.15 & 71 & 0.973 & 43 & 1.000 \\
 &  &  & 0.20 & 70 & 0.959 & 43 & 1.000 \\
 &  &  & 0.25 & 70 & 0.959 & 43 & 1.000 \\
 &  &  & 0.30 & 70 & 0.959 & 43 & 1.000 \\
 \hline
 &  & \multirow[t]{5}{*}{0.5} & 0.10 & 72 & 0.986 & 43 & 1.000 \\
 &  &  & 0.15 & 71 & 0.973 & 43 & 1.000 \\
 &  &  & 0.20 & 69 & 0.945 & 42 & 0.977 \\
 &  &  & 0.25 & 68 & 0.932 & 41 & 0.953 \\
 &  &  & 0.30 & 67 & 0.918 & 40 & 0.930 \\
 \hline
 &  & \multirow[t]{5}{*}{0.8} & 0.10 & 72 & 0.986 & 43 & 1.000 \\
 &  &  & 0.15 & 71 & 0.973 & 43 & 1.000 \\
 &  &  & 0.20 & 69 & 0.945 & 42 & 0.977 \\
 &  &  & 0.25 & 68 & 0.932 & 41 & 0.953 \\
 &  &  & 0.30 & 67 & 0.918 & 40 & 0.930 \\
\cline{1-8} \cline{2-8}
\multirow[t]{15}{3cm}{Type 2 Diabetes (All of Us)} & \multirow[t]{15}{*}{60} & \multirow[t]{5}{*}{0.2} & 0.10 & 54 & 0.900 & 34 & 0.971 \\
 &  &  & 0.15 & 54 & 0.900 & 34 & 0.971 \\
 &  &  & 0.20 & 52 & 0.867 & 33 & 0.943 \\
 &  &  & 0.25 & 52 & 0.867 & 33 & 0.943 \\
 &  &  & 0.30 & 52 & 0.867 & 33 & 0.943 \\
 \hline
 &  & \multirow[t]{5}{*}{0.5} & 0.10 & 54 & 0.900 & 34 & 0.971 \\
 &  &  & 0.15 & 54 & 0.900 & 34 & 0.971 \\
 &  &  & 0.20 & 52 & 0.867 & 33 & 0.943 \\
 &  &  & 0.25 & 52 & 0.867 & 33 & 0.943 \\
 &  &  & 0.30 & 50 & 0.833 & 31 & 0.886 \\
 \hline
 &  & \multirow[t]{5}{*}{0.8} & 0.10 & 54 & 0.900 & 34 & 0.971 \\
 &  &  & 0.15 & 54 & 0.900 & 34 & 0.971 \\
 &  &  & 0.20 & 52 & 0.867 & 33 & 0.943 \\
 &  &  & 0.25 & 52 & 0.867 & 33 & 0.943 \\
 &  &  & 0.30 & 50 & 0.833 & 31 & 0.886 \\
\cline{1-8} \cline{2-8}
\multirow[t]{15}{3cm}{Chronic Kidney Disease (All of Us)} & \multirow[t]{15}{*}{67} & \multirow[t]{5}{*}{0.2} & 0.10 & 65 & 0.970 & 40 & 1.000 \\
 &  &  & 0.15 & 65 & 0.970 & 40 & 1.000 \\
 &  &  & 0.20 & 64 & 0.955 & 39 & 0.975 \\
 &  &  & 0.25 & 64 & 0.955 & 39 & 0.975 \\
 &  &  & 0.30 & 64 & 0.955 & 39 & 0.975 \\
 \hline
 &  & \multirow[t]{5}{*}{0.5} & 0.10 & 65 & 0.970 & 40 & 1.000 \\
 &  &  & 0.15 & 65 & 0.970 & 40 & 1.000 \\
 &  &  & 0.20 & 64 & 0.955 & 39 & 0.975 \\
 &  &  & 0.25 & 64 & 0.955 & 39 & 0.975 \\
 &  &  & 0.30 & 62 & 0.925 & 37 & 0.925 \\
 \hline
 &  & \multirow[t]{5}{*}{0.8} & 0.10 & 65 & 0.970 & 40 & 1.000 \\
 &  &  & 0.15 & 65 & 0.970 & 40 & 1.000 \\
 &  &  & 0.20 & 64 & 0.955 & 39 & 0.975 \\
 &  &  & 0.25 & 64 & 0.955 & 39 & 0.975 \\
 &  &  & 0.30 & 62 & 0.925 & 37 & 0.925 \\
\cline{1-8} \cline{2-8}
\bottomrule
\end{longtable}

\begin{figure}
    \centering
    \includegraphics[width=1\linewidth]{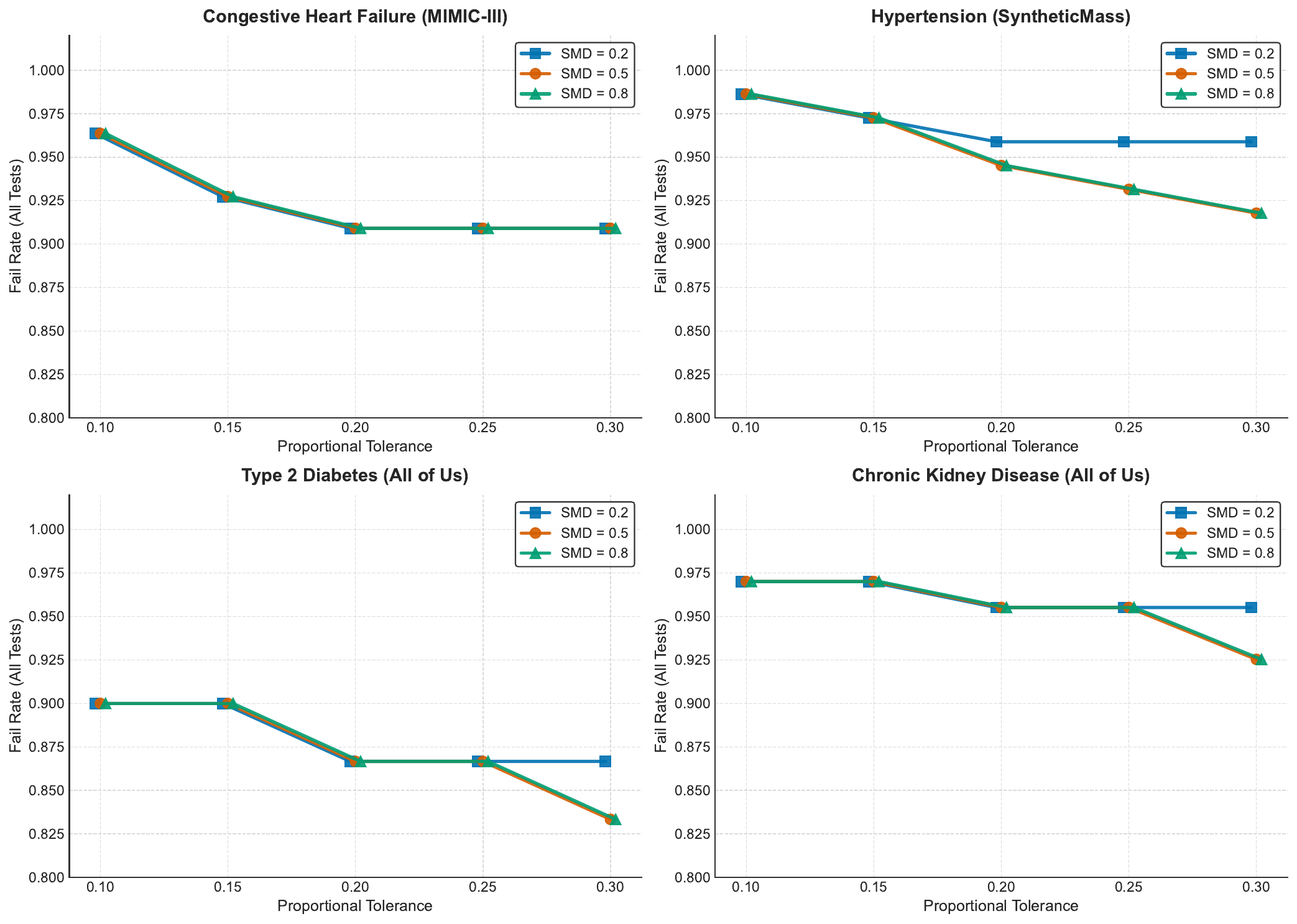}
    \caption{\textbf{Failure rates across statistical threshold combinations for all generated tests.} Failure rates for each cohort are shown across proportional tolerance thresholds of 10\%, 15\%, 20\%, 25\%, and 30\%, with separate curves for SMD cutoffs of 0.2, 0.5, and 0.8, considering all generated tests.}
    \label{fig:sensitivity_analysis_all}
\end{figure}
\FloatBarrier

\end{document}